\documentclass[11pt]{article}

\usepackage[final]{acl}

\usepackage{times}
\usepackage{latexsym}
\usepackage{amsthm}
\usepackage{amsmath}
\usepackage{booktabs}
\usepackage{xcolor}
\usepackage{amssymb}
\usepackage{array}
\usepackage{pifont}
\usepackage{capt-of,etoolbox}
\usepackage{multirow}
\usepackage{hhline}
\usepackage{bbding}
\usepackage{algorithm}
\usepackage{algorithmic}
\usepackage{boldline}
\usepackage{braket}
\usepackage{enumitem}
\usepackage{wrapfig,lipsum}
\usepackage{colortbl}
\usepackage{relsize}
\usepackage{ragged2e}
\usepackage{listings}
\usepackage{boldline}
\usepackage{arydshln}
\usepackage[T1]{fontenc}

\usepackage[utf8]{inputenc}

\usepackage{microtype}

\usepackage{inconsolata}

\usepackage{graphicx}

\title{When Does Defendant Statement Matter? A Study of Bias and Persuasion in LLM-Simulated Jurors}

\author{
  Cho-Ying Wu \\
  Bosch AI Research \\
  \texttt{Cho-Ying.Wu@us.bosch.com}
}

\begin{document}
\maketitle
\begin{abstract}
LLMs have been used to simulate human decision-making in professional settings, yet their behaviors in common-law jury trials remain unexplored. We study when and how a defendant’s courtroom statement affects LLM-simulated jurors, focusing on persuasion, ideological bias, and background-based affinity. To support the analysis, we introduce JuryBench, a benchmark containing controversial criminal cases in U.S. criminal law.
In each case, a defendant can claim various plausible justifications to support acquittal or reduced liability. 
We fix the base case and design defendants of different backgrounds, who give courtroom statements with varying emotional appeal or rebuttal. 
Jurors with diverse ideological profiles across the spectrum are simulated.
We examine 20 frontier LLMs, resulting in a total of 432K decisions and rationales, and quantify changes in verdict severity.
Our findings show that LLM-jury simulation echoes many human-jury findings. 
First, emotional persuasion can be detrimental, since jurors may perceive it as evidence of guilt or inconsistency.
Next, we show that background fit between jurors and defendants is a stronger and significant factor than other isolated factors, and that jurors are in general harsher toward opposite-background defendants and lenient toward same-background ones.  
Finally, we find that juror ideology also strongly shapes severity judgments.
These findings highlight both the promise and risks of using LLMs to model jury reasoning and call for careful evaluation.\footnote{The data and code are available at \url{https://github.com/choyingw/JuryBench}}
\end{abstract}

\begin{figure}[h]
    \centering
    \includegraphics[width=1.0\linewidth]{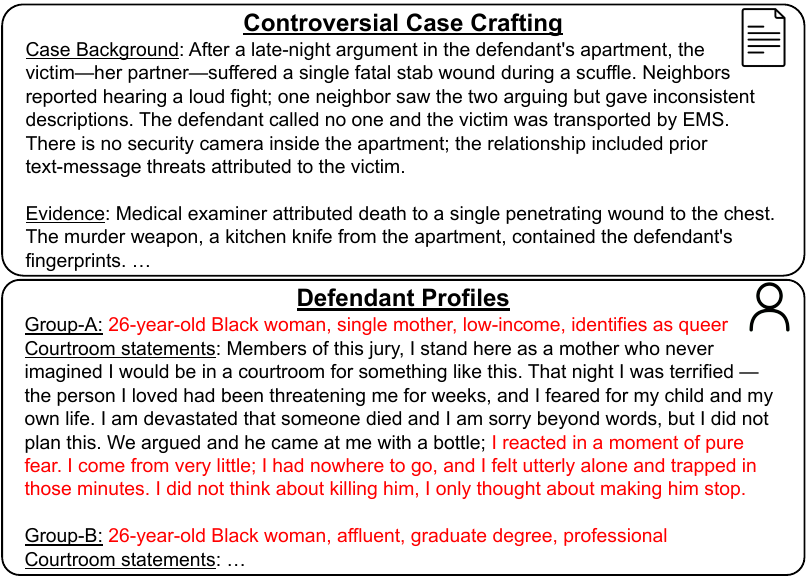}
    \includegraphics[width=1.0\linewidth]{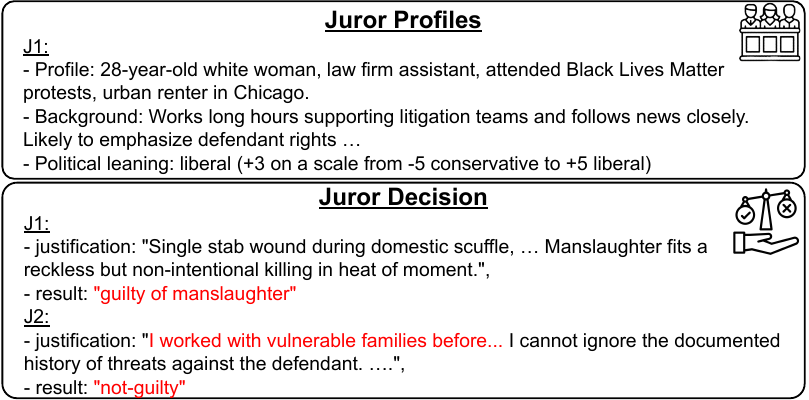}
    \vspace{-22pt}
    \caption{Our framework crafts controversial case scenarios and simulates defendant and juror profiles, to study how the LLM-jury makes decisions, especially in response to emotional persuasion.}
    \vspace{-6pt}
    
    \label{fig:main_effects_heatmap}
\end{figure}

\section{Introduction}
Large language models (LLMs) have shown strong ability to simulate interaction in professional scenarios, such as doctors and patients~\cite{kyungpatientsim, du2025llms, almansoori2025self, fan2025ai}, classroom scenarios ~\cite{zhang2025simulating,sanyal2025investigating,mannekote2025can}, or mental therapy~\cite{iftikhar2025llm}. In juristic scenarios, LLMs have been used to simulate courtroom debate and judgment ~\cite{shengbinyue2025multi, almansoori2025self} under the civil-law system. 
However, for the common-law system, a fundamental difference is that a \textbf{jury} is used in most criminal cases (Sec.~\ref{app:common-civil-law} for detailed comparison). 
A jury is composed of twelve people, selected from the public without legal training, whose function is to \textit{determine the facts and verdict}. This includes listening to the case details, understanding the defendant's background, evaluating the credibility of the evidence or witness, listening to the testimony, including the defendant's statement in response to examination, and finally deciding whether the defendant is guilty of a crime; while which crime is charged and sentences are decided by the prosecutor and judge. 
For example, a defendant claimed he only intended to beat the victim and did not intend to beat him to death. The jury will evaluate the evidence, like whether the injury is to a vital part, and decide whether the defendant is guilty of manslaughter.

More importantly, evidence and witnesses' credibility, reliability of statements from the defendant, and whether the defendant has valid grounds for \textit{justification}~\footnote{\textit{Justification} is a type of defense that exempts the defendant from liability, such as ~\textit{self-defense, act under duress or coercion, necessity, etc.} The jury needs to decide whether the justification is applicable. For example, a person who broke into a store because one was being chased may claim necessity, but the jury may not be convinced because it is not clear if one is under immediate danger. 
}, are evaluated based on a jury's commonsense, life experience, moral conviction, or even ideology~\cite{anwar2019politics}. The final verdict is voted from all jurors and requires unanimous consensus; if not, the judge will declare a mistrial (not acquittal), and the prosecutor needs to find more evidence or ask to substitute jurors.

In the common-law system, the jury selection is highly influential. 
Since the jurors lack legal training, the courtroom argument differs from the civil-law system that focuses on the applicability and interpretation of legal rules. Under the common-law system, the focus shifts to storytelling and emotional persuasion of the jury, where the jury's bias inevitably exists and requires careful study. Note that higher emotional contagion may help persuade the jury~\cite{erickson1978speech,bornstein2011jury}, but excessive rhetoric or performative emotional appeal can have negative effects~\cite{cramer2009expert,choi2023influence}.

To strategically win the trial, both the prosecutor and the defendant's lawyer can raise \textit{challenges}\footnote{Challenge of jury selection is a legal right and routinely used in trials. A lawyer still needs to explain detailed reasons, which are further reviewed by a judge, to reject jurors in most cases, and the reasons cannot be based on demographic features such as ethnicity or gender, or simply unsupported LLM discoveries.} to replace some jurors after background checks who might have adversarial bias against them.
Lawyers nowadays usually form a \textbf{focus group} before a trial, recruiting people with backgrounds similar to the jury and mock-trialing the case beforehand to understand potential bias. 
The process helps the defendant's lawyer develop new strategies or seek juror substitution. 
Some lawyers in practice start using LLM-jury to mock trials, revise trial strategies, and more importantly, \textit{reshape the statement and message brought to the jury that may emotionally move the jurors.} However, there is no prior research studying LLMs' ability to play the juror's role, including what the behavior of each frontier LLM is, how the LLM-jury reacts to the emotionally persuasive defendant statements, what factors affect each LLM the most, how could each LLM simulate the juror with different ideologies, and whether the decision reflects bias linked to the defendant and jury's background.   
This work presents the first systematic study to analyze LLM-jury and respond to those questions.

We present \textbf{JuryBench} that crafts 500 highly controversial cases under the US Criminal Code, where the defendants claim various plausible justifications or other reasons for their actions to be acquitted or face less severe charges. 
For each case, we fix the case background and given evidence, but we design different defendants and backgrounds, such as one defendant's background conforming to common stereotypes and the other's subverting them.    
Twelve jurors of different ideologies are designed and analyzed, including their decisions and reasons for each case.
A total of 20 frontier LLMs are examined for simulation, amounting to 432K decisions and reasons.

The contributions are summarized as follows.
\begin{itemize}
\item We present the first study on LLM-jury with focus specifically on emotional persuasion as the core dynamics in the common law. We study jury's bias and conduct analysis to understand the decisions and rationales behind.
\item We present JuryBench: A benchmark that includes 500 highly controversial criminal cases, around 400 potential charges, various defendant and juror backgrounds, and long-form defendant statements intended to support acquittal or reduced liability.   
\item We extensively evaluate 20 frontier LLMs, producing 432K juror decisions with reasons, to analyze their behaviors and relate the findings to legal psychology.


\end{itemize}

\section{Related Work}

\subsection{AI for the Juristic Domain}
Early developments of AI in the juristic domain include legal judgment prediction, which predicts what crimes and length of sentence a defendant will be charged with based on a judgment document~\cite{masala2021jurbert, han2026lawshift, hwang2022multi, trautmann2022legal,liu2023ml,strickson2020legal,feng2022legal,xiao2018cail2018,hu2025llms}, legal language understanding and reasoning to study how models read, interpret, and reason over legal texts ~\cite{chalkidis2022lexglue,fei2024lawbench,guha2023legalbench,han2025courtreasoner,chlapanis2025greekbarbench,akarajaradwong2025nitibench}, legal QA and retrieval that find the relevant legal authorities or evidence to support the question answering. ~\cite{louis2024interpretable,abdallah2023exploring,ryu2023retrieval,li2023sailer,dai2025laiw,shengbinyue2025multi}. 

Though the works have contributed to individual tasks, they did not adopt the LLM agents' strength for court simulation. There are two works closer to ours. AgentsCourt~\cite{he2024agentscourt} simulates courtroom debates and the final judgment. However, the debates are built upon judgment documents,
where judges have already reached a decision and written the case facts in a concise and conclusive manner. As a result, the defendant has limited room to make substantive arguments, and their statements reduce to repeatedly expressing remorse for several rounds, such as “once again, I am deeply regretful for ...” in many cases; while we focus on highly controversial scenarios, where the defendants can make effective claims. 
Another work, AgentCourt~\cite{chen2025agentcourt}, focuses on the application and interpretation of legal rules, but it does not examine how conflicting legal interests concerning the defendant are balanced through the jury's consideration.
More importantly, both works are under the civil-law system, where courtroom arguments naturally revolve around legal provisions that require dense juristic language.
Our work focuses on the common-law system, where lay jurors evaluate facts and verdicts, and
the courtroom arguments often appeal to commonsense, persuasion, and emotion. The work differs substantially from prior works as a pioneering research on the track.

\subsection{Human Jury Bias}
\label{sec:human_bias}
Jury bias has been widely studied in the legal-psychology domain. Some works point out that bias forms at a very early stage, during opening remarks or when jurors learn the defendant's background~\cite{kramer1990pretrial,kalven1966american}, and follow-up works note that jurors might seek to reinforce their constructed story from the first impression. 
The bias may concern the affinity between a juror’s and a defendant’s background. A juror may show mercy to someone similar to oneself and potentially judge outsiders more harshly~\cite{kerr1995defendant,rhodes2025partisan,foresta2025beyond}. However, a work also notes that in few cases the \textit{black sheep effects} exist, where jurors may judge a defendant who shares a similar background with the juror but violates the core values more harshly than an outsider~\cite{kerr1995defendant,marques1988black}.
Further, the bias may also concern ideology. Some studies show that conservative-attitude jurors are more likely to support a guilty verdict \cite{pyo2025mock,clark2012relationship,anwar2019politics} than liberal profiles. 

\subsection{LLM for Persona Simulation}
Persona simulation has been studied using LLMs for general conversational purposes~\cite{wang2025coser,hu2024quantifying,ni2026survey,wang2025beyond}, or specific domains such as doctor-patient~\cite{kyungpatientsim} and mental screening~\cite{wang2025talkdep}. The simulation does not explicitly instruct the models on what to say or how to react, but it implicitly sets an internal persona that guides the agent's responses, which may not explicitly reveal one's background. However, the persona simulation has not been used in the legal domain, especially to examine whether the LLM-jury can exhibit biases similar to those of humans. The work analyzes whether different personas affect trial outcomes.

\section{Simulation Framework}
To analyze LLM-jury, we require data containing detailed defendant background, base-case scenarios, courtroom speech, and each juror's background.
Specifically, we require controversial and debatable cases to understand a model's reasoning. 
The judgment documents usually include only base-case scenarios without other  materials and share the same concerns as prior works~\cite{he2024agentscourt} that the documents were written by judges after the verdict and sentences were determined. The facts are filtered and written in a conclusive manner, and they cannot reflect the case's uncertainty during the trial. Plus, juror profiles, ideology, and the defendant’s detailed backgrounds are not recorded or disclosed in public records for privacy and safety purposes (like US federal court for public access has restricted such information being released. Some information is purposely sealed and incomplete, especially for criminal cases to avoid retaliation.)

Instead, we use 
LLM and human experts to collaboratively craft the required materials.


\subsection{Case Generation}
\label{sec:cg}
We first use GPT-5.4 to generate 500 controversial and arguable criminal cases across around 400 potential criminal charges aligned with the US Criminal Code, and write the cases to .json files. Each entry includes "defendant background", "case background", and "evidence" that constitute a case. The case background and evidence describe the crime scene and found evidence to charge the defendant.
For defendant background, we first prompt to generate common stereotypes that may evoke empathy among people with a specific ideology, and form Group-A. Then we further generate labels that subvert the stereotypes, such as a defendant from a socially minor group is actually wealthy and successful, and they form Group-B. 

\textbf{Statement}.
For Group-A and Group-B defendants, based on the case background, evidence, and defendant profiles, we use GPT-5.4 to generate defendant statements during the examination by the lawyer or the prosecutor. The statements seek to emotionally persuade the jury by stressing their grounds for sympathy, justifying their conduct, expressing remorse, or combining these. Each statement contains around 15 sentences.

\textbf{Human in the loop}. The case and statement generation is manually reviewed by a practicing lawyer. We ask the expert to inspect the materials, including feasibility of the case background and evidence, ensuring the scenarios are controversial so defendants can make effective appeals while ensuring the statements conform to the base case scenario. 
Further, we ask the expert to rate the strength of emotional contagion $p$ and remorse $q$ for each statement on a scale of [1, 5], where 5 is the most powerful. 
For each defendant background, we ask the model to generate an affinity score \(f_a \in [-5, 5]\), indicating whether the background is more likely to elicit empathy from conservative- or liberal-attitude jurors. -5: the strongest conservative affinity; +5: the strongest liberal affinity.
The scores are checked by an expert trained in sociology.

\subsection{Juror Simulation}
We also use GPT-5.4 to generate each juror's profile and simulate various ideologies, following prior work on ideology simulation~\cite{argyle2023out,park2024generative}. Ideology scores \textbf{$f_{id}$} are rated from -5 (most conservative) to +5 (most liberal), and we ensure that the generated ideologies are uniformly distributed among all jurors.

\subsection{Case Decision}
We ask each simulated juror to decide each crafted case. The juror profiles are fed into an LLM as the system prompt, while the user prompt asks each juror to produce a decision $g$: not-guilty or guilty of a charge from a given list of potential charges.
This simulates real trials, where a judge instructs the jury on the charges available for consideration.
The decisions are made for all cases, including combinations of Group-A/B defendants and with/without statements. 
The user prompt instructs the LLM to consider all materials, including the defendant's statement if present, and focus on reasonable jury-style judgments. If the LLM cannot uphold a guilty decision, or if the evidence does not support the charge, return not guilty.
We ask the LLM to provide brief reasons for their decisions, and we give another pass to the model with prompts to check if the verdicts are consistent with those reasons. If not, regenerate the decisions.
 
Note that a case may involve multiple crimes; for example, assault and manslaughter may coexist, and we ask the LLM to output the most severe crime a defendant is guilty of. 
Further, the work distinguishes potential charges by \textit{mens rea}, i.e., intentional, reckless, or negligence when committing a crime, where the severity and sentence vary.   
We also consider attempted offenses, such as attempted murder or attempted arson, where the criminal act was not completed but still results in legal penalties. 
Prior common-law legal judgment prediction benchmarks did not simulate the jury and predict their decisions~\cite{sesodia2025annocaselaw,guha2023legalbench,semo2022classactionprediction}, or even consider the mens rea or attempted offenses. 


\textbf{Severity}. To quantitatively analyze the effect after the statement, we ask the human expert to label the severity $s$ of each potential charge and build a large dictionary for the crime severity. $s \in$ [0, 15], where 0 means not guilty or having valid justifications like under duress or complete self-defense. 15 is the heaviest. <= 5 is a misdemeanor, and > 5 is a felony with sentences typically more than one year. 

\section{Analysis and Findings}
We aim to answer the following questions through the analysis. 
Following prior works using general-domain LLMs for legal benchmarks~\cite{fan2025lexam,hu2025llms},
we examine 20 different frontier LLMs, assess their ability to simulate jurors, examine how the juror profile may affect the decision, and identify the underlying bias for each LLM. The temperature is set to 0.2 for generation with better consistency in the legal judgment if its API permits.
Though some legal-domain LLMs exist, they are mostly trained on Chinese civil-law cases with much smaller model sizes, such as LegalOne-R1 or DISC-LawLLM.

\noindent\textbf{Controversiality of crafted cases.} 
To verify if the crafted cases contain enough uncertainty so that the defendants can make effective statements for rebuttal,
we first evaluate how diverse the predictions are for a case. 
For each case, there are 960 = 20 (\# of models) $\times$ 12 (\# of jurors) $\times$ 4 (Group-A/B, with statement or not) predictions. We take the most frequent criminal charge and divide its count by 960 to obtain \textit{mode share}. The mode share averaged across 500 cases is 0.645, and the histogram peak is at 50\%-60\% (10\% per bucket). Only 3 cases have unanimous decisions. The statistics show the cases are highly controversial.    

\subsection{How do statements affect jury decisions?}
\label{sec:se}
For a case involving defendant $i$ and juror $j$, $s_{i,j}$ shows the severity score associated with the juror's decision. Then we compute statement effect, which is the severity change after the statement.   
\begin{equation}
\text{SE}_{i,j}
=
{s}^{\text{no-statement}}_{i,j}
-
{s}^{\text{statement}}_{i,j}.
\end{equation}
We average over the number of instances $N$ to get the average SE for a model. SE$>$0 means statements reduce verdict severity. We also compute success rate (SR), hurt rate (HR), and no-change rate (NR).

\begin{equation}
\text{SR}
=
\frac{
\sum_{i,j} \mathbf{1}\!\left(\text{SE}_{i,j} > 0\right)
}{
N
}.
\end{equation}

\begin{equation}
\text{HR}
=
\frac{
\sum_{i,j} \mathbf{1}\!\left(\text{SE}_{i,j} < 0\right)
}{
N
},
\end{equation}
where $\mathbf{1}$ is the indicator function that evaluates to 1 if the condition holds, and 0 otherwise. NR is simply 1$-$SR$-$HR.

\begin{table}[t]
\small
\centering
\vspace{-3pt}
\begin{tabular}{lcccc}
\hline
LLM & SE & SR & HR & NR\\
\hline
Gemini 3 Flash & 0.5492 & 0.1498 & 0.0532 & 0.7970\\
GPT-5 Mini & 0.3270 & 0.1413 & 0.0912 & 0.7675\\
Grok-4.3 & 0.3052 & 0.1865 & 0.1221 & 0.6914\\
Llama 4 & 0.0899 & 0.1539 & 0.1314 & 0.7147\\
GPT-5.4 Mini & 0.0479 & 0.1157 & 0.1058 & 0.7785\\
Gemini 3.1 Pro & -0.0481 & 0.0744 & 0.0704 & 0.8552\\
GPT-5.4 & -0.0597 & 0.0825 & 0.0880 & 0.8295\\
Kimi K2 Instruct & -0.0900 & 0.1370 & 0.1401 & 0.7229\\
GPT-5.5 & -0.1067 & 0.0882 & 0.0895 & 0.8223\\
Kimi K2.5 & -0.1395 & 0.1586 & 0.1630 & 0.6784\\
GPT-4o Mini & -0.2009 & 0.0587 & 0.0834 & 0.8578\\
Kimi K2.6 & -0.2062 & 0.1455 & 0.1613 & 0.6933\\
GPT-5.4 Nano & -0.2251 & 0.1179 & 0.1358 & 0.7463\\
Claude Sonnet 4.6 & -0.2431 & 0.0772 & 0.1150 & 0.8078\\
Claude Haiku 4.5 & -0.2477 & 0.1181 & 0.1411 & 0.7408\\
DeepSeek V4 Pro & -0.2770 & 0.1179 & 0.1433 & 0.7338\\
DeepSeek V4 Flash & -0.3043 & 0.1088 & 0.1409 & 0.7502\\
GLM 5 & -0.3841 & 0.1231 & 0.1650 & 0.7119 \\
Claude Opus 4.6 & -0.3854 & 0.0788 & 0.1366 & 0.7847\\
Qwen3 & -0.5917 & 0.1117 & 0.1889 & 0.6994\\
\hline
\end{tabular}
\vspace{-3pt}
\caption{\textbf{Effect of defendant's statement}.}
\vspace{-9pt}
\label{tab:effect1}
\end{table}

Table~\ref{tab:effect1} shows the results. First, one can find that the LLM-jury may not be easily swayed by the statement with about 70\%-80\% NR. The observation matches a famous rule from empirical legal studies, which states that around 80\% of jurors make up their minds after hearing the opening remarks regarding case background and evidence~\cite{kalven1966american, polavin2022jurors}.
Some follow-up studies show that jurors may construct a "story" as early as possible and seek evidence to support it~\cite{schweitzer2021effect}. A recent experiment shows that around 66\% to 75\% of jurors did not change opinions from their first impression~\cite{polavin2022jurors}. Our NR aligns with those studies, showing LLM-jury may exhibit a similar form of decision inertia and not easily change opinions once the case background is given.

Next, in line with SE, SR, and HR, more models tend to be harsher than lenient.
The observation echoes many legal-psychological studies~\cite{salekin1995influencing,van2025emotional,corwin2012defendant,proeve2023addressing} that note defendant statements can be a double-edged sword, not always leading to mitigation. The statements may positively affect the jury and reduce the severity. For example, in a case in JuryBench, the defendant is charged with misprision of a murder weapon, and from the defendant's words, she wanted to protect her family from retaliation. The same Gemini-3-Flash juror who made a guilty decision without the statement then recognizes this defense and makes a not-guilty decision. 
Yet, Some jurors may interpret highly intense emotions or incongruent statements as signals of guilt. For example, some juror reasons by Claude Sonnet-4.6 mention "The inconsistent statements further weaken credibility" with guilty decisions. Without the statements, the same jurors would have given not-guilty verdicts. 

From another perspective, the defendant statements may imply remorse, which jurors may find relatable and thereby reduce the severity, but some jurors may further reinforce the belief that the defendant is responsible because the expressed remorse is interpreted as an implicit admission. 
In one example, a GPT-4o-Mini juror states, "While there was regret, the defendant's position of privilege and manner do not evoke the same level of sympathy ..."
Without the statement, the same juror gave a not-guilty verdict.




\subsection{Which latent factors affect jury decisions?}
\label{sec:q2}

\begin{figure}[h]
    \centering
    \includegraphics[width=0.95\linewidth]{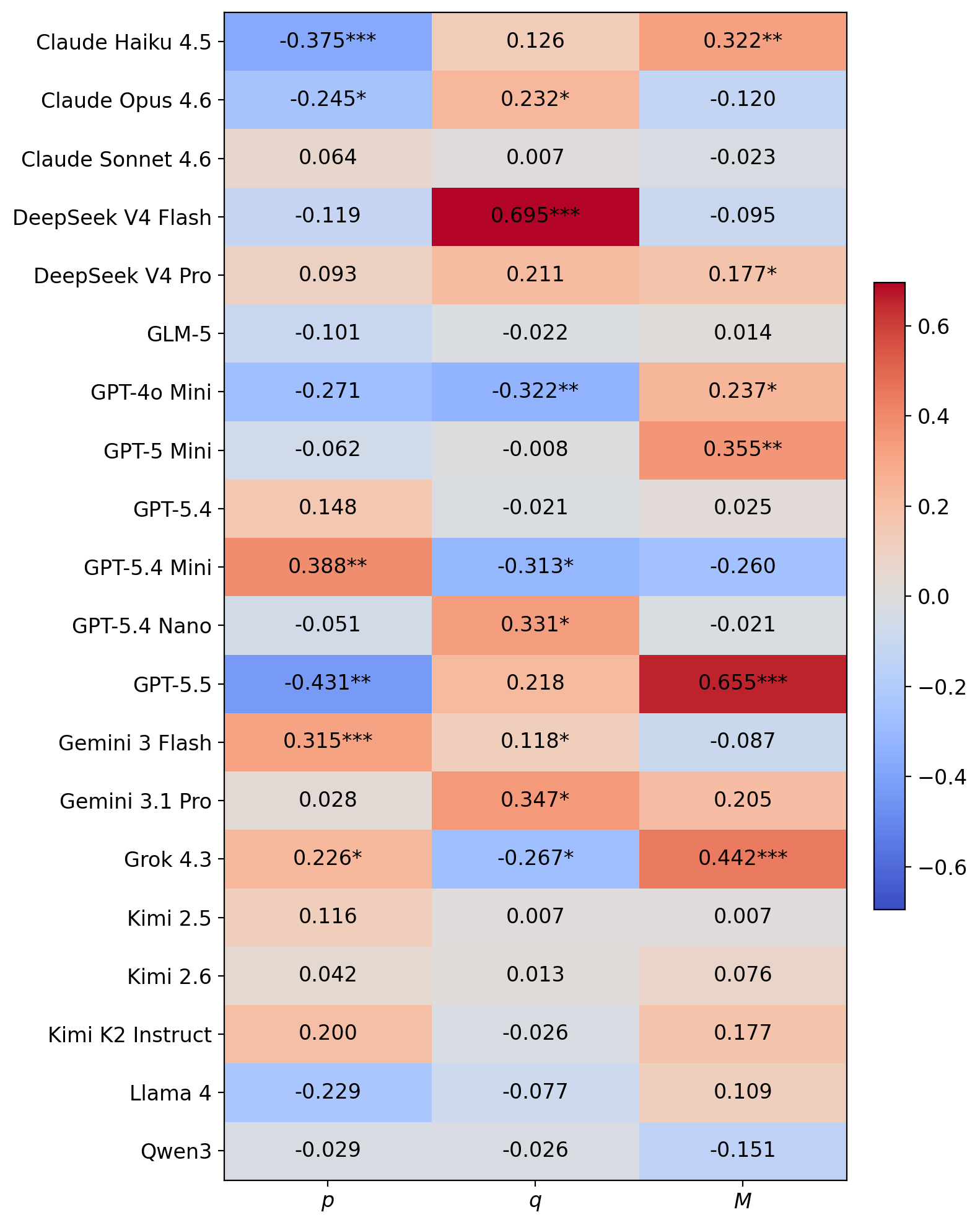}
    \vspace{-11pt}
    \caption{\textbf{Main Effect Analysis.} Heatmap for the coefficients are shown. * indicates significance based on statistical p-value < 0.05; ** p < 0.01; *** p < 0.001.}
    \vspace{-5pt}
    \label{fig:main_effects_heatmap}
\end{figure}

To answer the question, we identify three factors: emotional contagion $p$, remorse $q$, and background fit: for a defendant $i$ and a juror $j$, the background fit $M^{i,j}$ is computed by the product of the defendant's affinity and the juror's ideology. 
\begin{equation}
M^{i,j} = f_{id}^j \cdot f_{a}^i,
\end{equation}
where larger positive values show higher match, and more negative values show stronger mismatch.  
To decide which factor is more important when severity scores change, we build a model
\begin{equation}
\text{SE}=\alpha+\beta_1z(p)+\beta_2z(q)+\beta_3z(M),
\label{linear_eq}
\end{equation}
where $z()$ denotes the standardization to zero mean and unit variance. To analyze which factor contributes most, we calculate these coefficients via fitting the linear model by least squares over all defendant and juror instances. The heatmap plot for each term's coefficient is shown in Fig.~\ref{fig:main_effects_heatmap}. 
We compute statistical p-values for each coefficient and mark statistical significance with three thresholds.

\textbf{Emotional Contagion}.
Higher emotional contagion reduces severity for a modest majority of models.
Some LLMs are significantly affected by emotional contagion, such as Gemini-3-Flash, and we have given an example in Sec.~\ref{sec:se}. 
In contrast, it has negative effects for some models, such as Claude-Haiku-4.5. 
For example, in a case related to bribery, a Claude-Haiku juror thinks, "The defendant’s charismatic speech and appeals to service do not overcome the financial and transactional evidence and his own admission." 
Without the statement, the juror reaches a not-guilty verdict. The rationale resembles the human mindset, where higher rhetoric may risk distrust and prompt closer scrutiny of the claim with evidence.

\textbf{Remorse}.
Expressing remorse reduces severity for a sizable minority of models.
DeepSeek-V4, Gemini, and some GPT models emphasize this factor. 
For example, DeepSeek models often state "The defendant's speech is remorseful and logical" in the juror rationales and subsequently reduce the severity.
In contrast, Grok and a few GPT models exhibit the opposite effect- showing remorse may actually reinforce perceptions of guilt, where "Though feeling regretful, he did admit ..." are mentioned in many juror rationales. For half of the examined models, remorse has neutral effects. 

\begin{figure*}[h]
    \centering
    \includegraphics[width=0.90\linewidth]{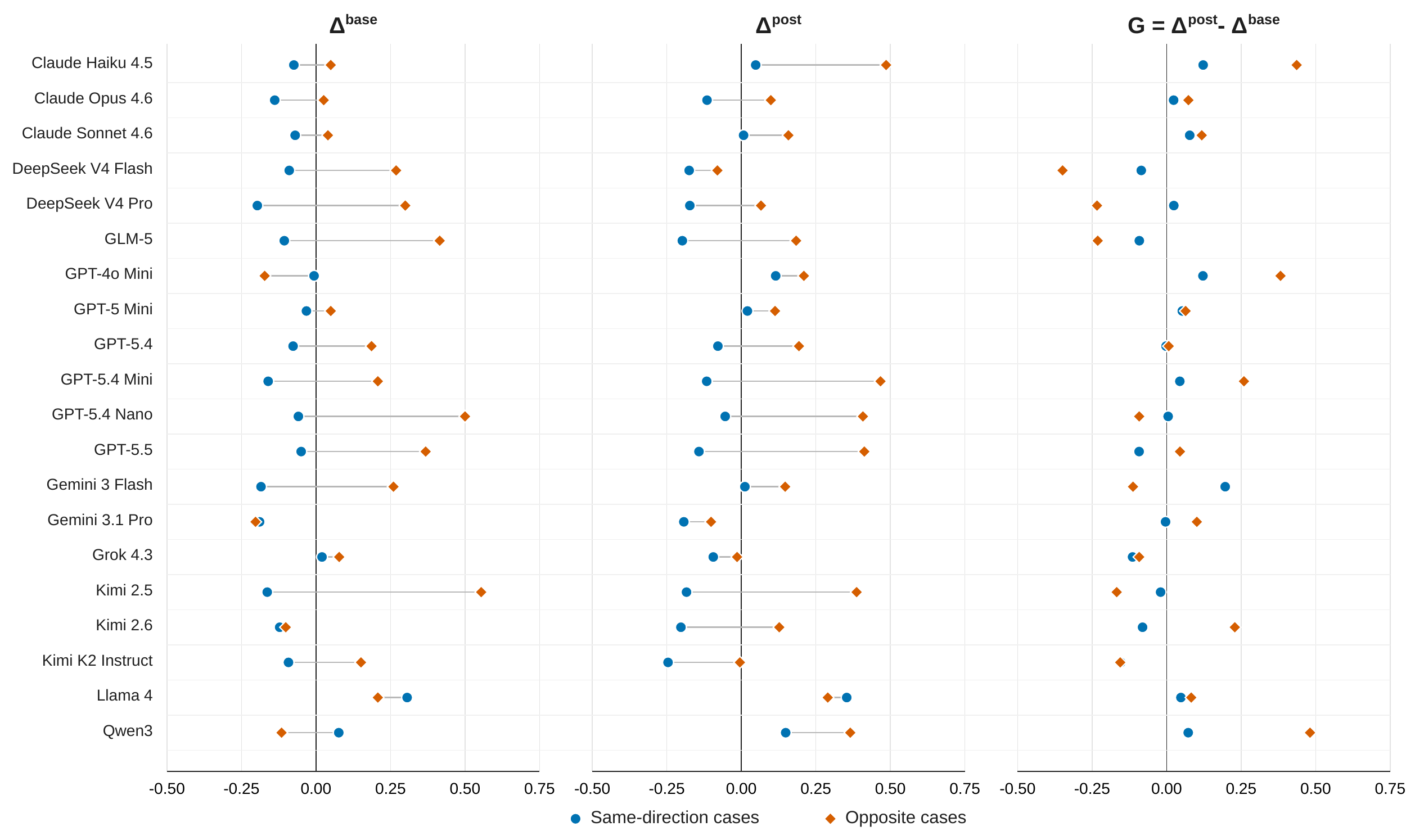}
    \vspace{-7pt}
    \caption{\textbf{Analysis for the matched (same-direction) and mismatched (opposite) background fit.}}
    \vspace{-3pt}
    \label{fig:same_vs_opposite}
\end{figure*}

\textbf{Background Fit}.
Surprisingly, the background fit is a more significant factor than emotional contagion and remorse, with a substantial majority of models showing positive coefficients with significance, and none of the negative coefficients are statistically significant.
GPT-5.5, Grok-4.3, GPT-5-mini, and Claude-Haiku-4.5 are the most prominent. 
For instance, a GPT-5-mini's juror said
"As a mother I feel deep compassion, and the defendant’s speech about fear for her baby and panic is credible", or "As a father and small-business owner I trust the defendant..."
The jurors sometimes relate the defendant's background to their own and reduce the severity.

Likewise, if a defendant's background is the opposite of a juror’s, one's statement might have negative effects. For example, in one case the defendant said, "I have run a business for thirty years and I have paid my taxes and raised my children to respect the law"; however, a juror with the opposite ideology escalates the severity after hearing the statement and said, "His self-defense claim lacks corroborating evidence, and his speech focused heavily on his privileged status rather than genuine remorse ..."

The above analysis shows that LLM-simulated jurors exhibit patterns that echo
human-jury findings, implicitly persuaded by emotion, remorse, or background bias, but may also develop distrust as the legal-psychology studies suggest (Sec.~\ref{sec:human_bias}). In the appendix, we conduct human evaluation on a subset of the generated cases to support the findings of similarity between LLMs and human reasoning. We also consider interactions among the terms in Eq.~\ref{linear_eq} and analyze a conditional model beyond the marginal effects.



\begin{figure*}[h]
    \centering
    \includegraphics[width=0.91\linewidth]{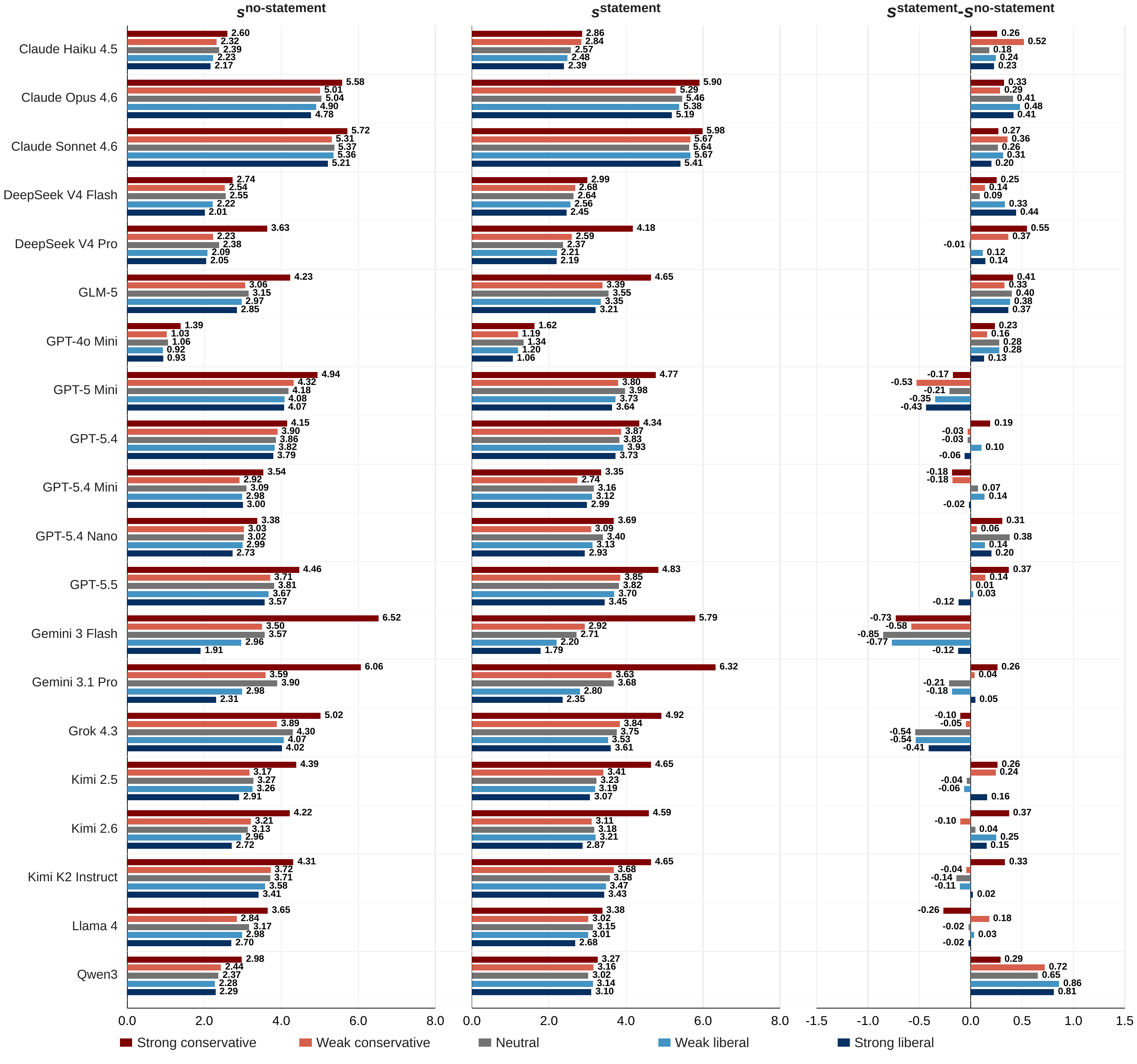}
    \vspace{-8pt}
    \caption{\textbf{Analysis for liberal/ conservative jury.}}
    \vspace{-3pt}
    \label{fig:liberal_conservative}
\end{figure*}

\subsection{Do Group-A and Group-B differ in verdict severity?}
In Sec.~\ref{sec:cg}, we generate two defendants for Group-A and Group-B for a case. The analysis here uses $i$ as the case index and $j$ as the juror index, and splits the defendants into $\{iA, iB\}\in i$. We define the estimators $\Delta^{\text{base}}$ and $\Delta^{\text{post}}$, which are the mean group difference over all instances before and after a statement.

\begin{equation}
\Delta^{\text{base}}_{i,j}
=
{s}^{\text{no-statement}}_{iA,j}
-
{s}^{\text{no-statement}}_{iB,j}, 
\end{equation}
\begin{equation}
\Delta^{\text{post}}_{i,j}
=
{s}^{\text{statement}}_{iA,j}
-
{s}^{\text{statement}}_{iB,j}, 
\end{equation}
where positive scores indicate Group-A has higher severity, and vice versa.
We also define signed gap of shift as 
\begin{equation}
\begin{aligned}
G_{i,j}
&= \Delta^{\text{post}}_{i,j}-\Delta^{\text{base}}_{i,j} \\
&\scriptstyle
=({s}^{\text{statement}}_{iA,j}-{s}^{\text{no-statement}}_{iA,j})
-({s}^{\text{statement}}_{iB,j}-{s}^{\text{no-statement}}_{iB,j}),
\end{aligned}
\end{equation}
where $G_{i,j}<0$ suggests Group-A becomes relatively less severe compared with Group-B, or Group-B becomes more severe relative to Group A, while $G_{i,j}>0$ is the opposite. $G$ is the mean over all instances.

To conduct the analysis, we separate cases by background fit $M^{i,j} > 0$ (same direction) and $M^{i,j} < 0$ (opposite direction). The results are shown in Fig.~\ref{fig:same_vs_opposite}. 

\textbf{Group-A v.s. Group-B}. Observing from $\Delta^{\text{base}}$ and $\Delta^{\text{post}}$, one can find that nearly all the models show a pattern that the same-direction cases have negative scores, and the opposite cases have positive scores.
The former indicates Group-A has lower severity than B on average, since more matched backgrounds lead to greater leniency from jurors of similar backgrounds, compared to Group-B that counters the stereotype. 
Conversely, the latter shows Group-B is less severe, since jurors with opposite backgrounds may criticize Group-A more with higher mismatch scores.


\textbf{Before vs. after statement.}
We further examine the signed gap of difference
$G$. 
For most of the models, the signs of $G$ for both the same-direction and opposite direction are the same,
suggesting that each 
model has its own systematic tendency to enlarge or close the gap of difference between Group-A and Group-B, after hearing the statements. 
Across the models, the signs are roughly evenly divided between positive and negative without a clear inclination.

The magnitude $|G|$ is often larger for opposite-direction cases. This indicates that opposite-direction cases experience stronger post-statement movement in the signed gap of difference, suggesting greater sensitivity to the statement intervention.

\subsection{How does ideology affect jury's decisions?}
We further examine how a juror’s ideology affects their decision. We divide the analysis into 5 buckets for each LLM based on the ideology score: strong conservative (-5 to -3), weak conservative (-2 to -1), neutral (0), weak liberal (1 to 2), strong liberal (3 to 5). 
Severity scores before and after the statements are used as the metrics. Fig.~\ref{fig:liberal_conservative} shows the results. 

As shown by $s^{\text{no-statement}}$ and $s^{\text{statement}}$,
an obvious trend across LLMs shows that conservative jurors assign higher severity scores than neutral jurors, who are also harsher than liberal jurors. Overall, liberal jurors are clearly more lenient than conservative jurors.
On average, stronger ideological leaning also indicates stronger effects- the strong conservatives always give the highest severity, and the strong liberals are usually the lowest. 

From $s^{\text{statement}}-s^{\text{no-statement}}$, around 60\% of LLMs show the same signs of severity change across all ideologies. This suggests that each model has its own tendency in how statements affect its severity judgments.
Then, around 70\% of LLMs show larger post-statement changes for conservative jurors than for the liberal, suggesting that conservative jurors are more sensitive to the statement.

Our studies echo prior legal-psychological research~\cite{pyo2025mock,sivasubramaniam2020jury,anwar2019politics}: conservative or “crime control” oriented jurors are more conviction-prone and tend to seek evidence to support a guilty verdict.


\section{Conclusion}
This work presents the first systematic study of LLM-jury under the common-law setting. We introduce JuryBench, a benchmark of controversial criminal cases, diverse juror profiles, controlled defendant backgrounds, and emotionally persuasive statements to enable fine-grained analysis of LLM-jury behavior. 
LLM-jury echoes several legal-psychology findings: sticky early impressions, double-edged defendant statements, and ideology- or affinity-driven decisions.
JuryBench provides a controlled testbed for studying how LLMs may support mock-jury analysis and for lawyers to conduct mock trials with a focus group to revise the trial strategies. The work further supports future work on transparent and accountable legal simulations.





\section*{Limitations}
The work has several limitations:
\begin{itemize}

\item The work presents a \textbf{simplified court simulation}. In real trials, the case background and evidence introduction are presented during the opening statements and the prosecution's case-in-chief, and we combine them into a single stage to provide a basic understanding. The work focuses on understanding the LLM-jury's reaction to emotional persuasion; therefore, we simplified the testimony and cross-examination to speech from the defendant, which is also the core dynamics of modern trials for emotional persuasion. The full simulation of cross-examination, such as on the credibility of other evidence, may include complex arguments, which is orthogonal to emotional persuasion and not within the work's scope. Last, the real verdict requires unanimous consensus from the jury in most U.S. states through group discussion. Since the focus of the work is to study how each LLM-simulated juror with different controlled parameters responds to emotional persuasion, the later group discussion to reach consensus is out of scope and we do not simulate the stage. 

\item The work focuses on U.S. common-law criminal trials. While our work examines the implicit factors shaping the decision-making process of LLM-jury, which is not limited to country or case type, the current scope primarily draws on U.S. criminal charges. 

\item The work only uses text to simulate the courtroom process following the previous work. However, the emotional contagion and remorse could be amplified by body language, facial expressions, and other factors that are only observable in real courtroom settings.

\item With fast iteration and an increasing number of LLMs coming out, we only chose 20 frontier LLMs for study. The relative behaviors observed across models may change as model providers update training data, safety policies, and inference systems in the newer versions.
\end{itemize}

\section*{Ethical Considerations}
The work is designed as a research tool to understand the behaviors of LLM-jury, especially provided to legal professionals who intend to conduct mock trials to plan or refine their trial strategies using LLMs and simulate jurors with similar profiles. 
The intent of the work is not to automate real legal judgment or interfere with real cases. 
The cases, defendants, and jurors in our benchmark are synthetic and do not describe real individuals; nevertheless, the generated profiles involve sensitive social attributes and criminal allegations, so they should be handled with care. 
We will include clear documentation, intended-use restrictions, and warnings that the data may contain stereotypical associations introduced intentionally for bias analysis in the benchmark release.

\section*{Acknowledgments}
Per ACL policy, we disclose the generative AI usage in paper. We use generative AI to check contents correctness and polish the writing, while the content originality and research ideas are fundamentally developed by humans.

\bibliography{custom}

\begin{thebibliography}{61}
\providecommand{\natexlab}[1]{#1}

\bibitem[{Abdallah et~al.(2023)Abdallah, Piryani, and Jatowt}]{abdallah2023exploring}
Abdelrahman Abdallah, Bhawna Piryani, and Adam Jatowt. 2023.
\newblock Exploring the state of the art in legal qa systems.
\newblock \emph{Journal of Big Data}, 10(1):127.

\bibitem[{Akarajaradwong et~al.(2025)Akarajaradwong, Pothavorn, Chaksangchaichot, Tasawong, Nopparatbundit, Pratai, and Nutanong}]{akarajaradwong2025nitibench}
Pawitsapak Akarajaradwong, Pirat Pothavorn, Chompakorn Chaksangchaichot, Panuthep Tasawong, Thitiwat Nopparatbundit, Keerakiat Pratai, and Sarana Nutanong. 2025.
\newblock Nitibench: Benchmarking llm frameworks on thai legal question answering capabilities.
\newblock In \emph{Proceedings of the 2025 Conference on Empirical Methods in Natural Language Processing}.

\bibitem[{Almansoori et~al.(2025)Almansoori, Kumar, and Cholakkal}]{almansoori2025self}
Mohammad Almansoori, Komal Kumar, and Hisham Cholakkal. 2025.
\newblock Self-evolving multi-agent simulations for realistic clinical interactions.
\newblock \emph{arXiv preprint arXiv:2503.22678}.

\bibitem[{Anwar et~al.(2019)Anwar, Bayer, and Hjalmarsson}]{anwar2019politics}
Shamena Anwar, Patrick Bayer, and Randi Hjalmarsson. 2019.
\newblock Politics in the courtroom: Political ideology and jury decision making.
\newblock \emph{Journal of the European Economic Association}, 17(3):834--875.

\bibitem[{Argyle et~al.(2023)Argyle, Busby, Fulda, Gubler, Rytting, and Wingate}]{argyle2023out}
Lisa~P Argyle, Ethan~C Busby, Nancy Fulda, Joshua~R Gubler, Christopher Rytting, and David Wingate. 2023.
\newblock Out of one, many: Using language models to simulate human samples.
\newblock \emph{Political Analysis}, 31(3):337--351.

\bibitem[{Bornstein and Greene(2011)}]{bornstein2011jury}
Brian~H Bornstein and Edie Greene. 2011.
\newblock Jury decision making: Implications for and from psychology.
\newblock \emph{Current directions in psychological science}, 20(1):63--67.

\bibitem[{Chalkidis et~al.(2022)Chalkidis, Jana, Hartung, Bommarito, Androutsopoulos, Katz, and Aletras}]{chalkidis2022lexglue}
Ilias Chalkidis, Abhik Jana, Dirk Hartung, Michael Bommarito, Ion Androutsopoulos, Daniel Katz, and Nikolaos Aletras. 2022.
\newblock Lexglue: A benchmark dataset for legal language understanding in english.
\newblock In \emph{Proceedings of the 60th Annual Meeting of the Association for Computational Linguistics (Volume 1: Long Papers)}, pages 4310--4330.

\bibitem[{Chen et~al.(2025)Chen, Fan, Gong, Xie, Li, Liu, Li, Qu, Alinejad{-}Rokny, Ni, and Yang}]{chen2025agentcourt}
Guhong Chen, Liyang Fan, Zihan Gong, Nan Xie, Zixuan Li, Ziqiang Liu, Chengming Li, Qiang Qu, Hamid Alinejad{-}Rokny, Shiwen Ni, and Min Yang. 2025.
\newblock Agentcourt: Simulating court with adversarial evolvable lawyer agents.
\newblock In \emph{Findings of the Association for Computational Linguistics: ACL 2025}, pages 5850--5865.

\bibitem[{Chlapanis et~al.(2025)Chlapanis, Galanis, Aletras, and Androutsopoulos}]{chlapanis2025greekbarbench}
Odysseas~S Chlapanis, Dimitrios Galanis, Nikolaos Aletras, and Ion Androutsopoulos. 2025.
\newblock Greekbarbench: A challenging benchmark for free-text legal reasoning and citations.
\newblock In \emph{EMNLP Findings}.

\bibitem[{Choi et~al.(2023)Choi, Nu{\~n}ez, and Wilkowski}]{choi2023influence}
Samuel Choi, Narina Nu{\~n}ez, and Benjamin~M Wilkowski. 2023.
\newblock The influence of attorney anger on juror decision making.
\newblock \emph{Psychiatry, Psychology and Law}, 30(3):271--298.

\bibitem[{Clark and Wink(2012)}]{clark2012relationship}
John~W Clark and Kenneth Wink. 2012.
\newblock The relationship between political ideology and punishment: What do jury panel members say?
\newblock \emph{Applied Psychology in Criminal Justice}, 8(2).

\bibitem[{Corwin et~al.(2012)Corwin, Cramer, Griffin, and Brodsky}]{corwin2012defendant}
Emily~P Corwin, Robert~J Cramer, Desiree~A Griffin, and Stanley~L Brodsky. 2012.
\newblock Defendant remorse, need for affect, and juror sentencing decisions.
\newblock \emph{Journal of the American Academy of Psychiatry and the Law Online}, 40(1):41--49.

\bibitem[{Cramer et~al.(2009)Cramer, Brodsky, and DeCoster}]{cramer2009expert}
Robert~J Cramer, Stanley~L Brodsky, and Jamie DeCoster. 2009.
\newblock Expert witness confidence and juror personality: Their impact on credibility and persuasion in the courtroom.
\newblock \emph{Journal of the American Academy of Psychiatry and the Law Online}, 37(1):63--74.

\bibitem[{Dai et~al.(2025)Dai, Feng, Huang, Jia, Xie, Zhang, Han, Tian, and Wang}]{dai2025laiw}
Yongfu Dai, Duanyu Feng, Jimin Huang, Haochen Jia, Qianqian Xie, Yifang Zhang, Weiguang Han, Wei Tian, and Hao Wang. 2025.
\newblock Laiw: A chinese legal large language models benchmark.
\newblock In \emph{Proceedings of the 31st International conference on computational linguistics}, pages 10738--10766.

\bibitem[{Du et~al.(2025)Du, Zheng, Hu, Xu, Li, Sun, Chen, Wu, Cai, and Ying}]{du2025llms}
Zhuoyun Du, Lujie Zheng, Renjun Hu, Yuyang Xu, Xiawei Li, Ying Sun, Wei Chen, Jian Wu, Haolei Cai, and Haochao Ying. 2025.
\newblock Llms can simulate standardized patients via agent coevolution.
\newblock In \emph{Proceedings of the 63rd Annual Meeting of the Association for Computational Linguistics (Volume 1: Long Papers)}, pages 17278--17306.

\bibitem[{Erickson et~al.(1978)Erickson, Lind, Johnson, and O'Barr}]{erickson1978speech}
Bonnie Erickson, E~Allan Lind, Bruce~C Johnson, and William~M O'Barr. 1978.
\newblock Speech style and impression formation in a court setting: The effects of “powerful” and “powerless” speech.
\newblock \emph{Journal of experimental social psychology}, 14(3):266--279.

\bibitem[{Fan et~al.(2026)Fan, Ni, Merane, Tian, Hermstr{\"u}wer, Huang, Akhtar, Salimbeni, Geering, Dreyer, Brunner, Leippold, Sachan, Stremitzer, Engel, Ash, and Niklaus}]{fan2025lexam}
Yu~Fan, Jingwei Ni, Jakob Merane, Yang Tian, Yoan Hermstr{\"u}wer, Yinya Huang, Mubashara Akhtar, Etienne Salimbeni, Florian Geering, Oliver Dreyer, Daniel Brunner, Markus Leippold, Mrinmaya Sachan, Alexander Stremitzer, Christoph Engel, Elliott Ash, and Joel Niklaus. 2026.
\newblock Lexam: Benchmarking legal reasoning on 340 law exams.
\newblock \emph{ICLR}.

\bibitem[{Fan et~al.(2025)Fan, Wei, Tang, Chen, Siyuan, Wei, and Huang}]{fan2025ai}
Zhihao Fan, Lai Wei, Jialong Tang, Wei Chen, Wang Siyuan, Zhongyu Wei, and Fei Huang. 2025.
\newblock Ai hospital: Benchmarking large language models in a multi-agent medical interaction simulator.
\newblock In \emph{Proceedings of the 31st International Conference on Computational Linguistics}, pages 10183--10213.

\bibitem[{Fei et~al.(2023)Fei, Shen, Zhu, Zhou, Han, Zhang, Chen, Shen, and Ge}]{fei2024lawbench}
Zhiwei Fei, Xiaoyu Shen, Dawei Zhu, Fengzhe Zhou, Zhuo Han, Songyang Zhang, Kai Chen, Zongwen Shen, and Jidong Ge. 2023.
\newblock Lawbench: Benchmarking legal knowledge of large language models.
\newblock \emph{arXiv preprint arXiv:2309.16289}.

\bibitem[{Feng et~al.(2022)Feng, Li, and Ng}]{feng2022legal}
Yi~Feng, Chuanyi Li, and Vincent Ng. 2022.
\newblock Legal judgment prediction via event extraction with constraints.
\newblock In \emph{Proceedings of the 60th annual meeting of the association for computational linguistics (volume 1: long papers)}, pages 648--664.

\bibitem[{Foresta(2025)}]{foresta2025beyond}
Alessandra Foresta. 2025.
\newblock Beyond a reasonable doubt: The impact of jurors’ political affiliations on trials: Evidence from north carolina.
\newblock \emph{The Journal of Law and Economics}, 68(2):361--386.

\bibitem[{Guha et~al.(2023)Guha, Nyarko, Ho, R{\'{e}}, Chilton, Aditya, Chohlas{-}Wood, Peters, Waldon, Rockmore, Zambrano, Talisman, Hoque, Surani, Fagan, Sarfaty, Dickinson, Porat, Hegland, Wu, Nudell, Niklaus, Nay, Choi, Tobia, Hagan, Ma, Livermore, Rasumov{-}Rahe, Holzenberger, Kolt, Henderson, Rehaag, Goel, Gao, Williams, Gandhi, Zur, Iyer, and Li}]{guha2023legalbench}
Neel Guha, Julian Nyarko, Daniel~E. Ho, Christopher R{\'{e}}, Adam Chilton, K.~Aditya, Alex Chohlas{-}Wood, Austin Peters, Brandon Waldon, Daniel~N. Rockmore, Diego Zambrano, Dmitry Talisman, Enam Hoque, Faiz Surani, Frank Fagan, Galit Sarfaty, Gregory~M. Dickinson, Haggai Porat, Jason Hegland, and 21 others. 2023.
\newblock Legalbench: A collaboratively built benchmark for measuring legal reasoning in large language models.
\newblock \emph{Advances in neural information processing systems}, 36:44123--44279.

\bibitem[{Han et~al.(2025{\natexlab{a}})Han, Takashima, Shen, Liu, Liu, Thuo, Knowlton, Piskac, Shapiro, and Cohan}]{han2025courtreasoner}
Sophia~Simeng Han, Yoshiki Takashima, Shannon~Zejiang Shen, Chen Liu, Yixin Liu, Roque~K Thuo, Sonia Knowlton, Ruzica Piskac, Scott~J Shapiro, and Arman Cohan. 2025{\natexlab{a}}.
\newblock Courtreasoner: Can llm agents reason like judges?
\newblock In \emph{Proceedings of the 2025 Conference on Empirical Methods in Natural Language Processing}.

\bibitem[{Han et~al.(2025{\natexlab{b}})Han, Yang, Feng, Huang, Xuxing, Li, Ge, and Ng}]{han2026lawshift}
Zhuo Han, Yi~Yang, Yi~Feng, Wanhong Huang, Ding Xuxing, Chuanyi Li, Jidong Ge, and Vincent Ng. 2025{\natexlab{b}}.
\newblock Lawshift: Benchmarking legal judgment prediction under statute shifts.
\newblock \emph{Advances in Neural Information Processing Systems Datasets and Benchmarks Track}.

\bibitem[{He et~al.(2024)He, Cao, Wang, Jin, Chen, Xu, Li, Liu, and Zhao}]{he2024agentscourt}
Zhitao He, Pengfei Cao, Chenhao Wang, Zhuoran Jin, Yubo Chen, Jiexin Xu, Huaijun Li, Kang Liu, and Jun Zhao. 2024.
\newblock Agentscourt: Building judicial decision-making agents with court debate simulation and legal knowledge augmentation.
\newblock In \emph{Findings of the Association for Computational Linguistics: EMNLP 2024}, pages 9399--9416.

\bibitem[{Hu and Collier(2024)}]{hu2024quantifying}
Tiancheng Hu and Nigel Collier. 2024.
\newblock Quantifying the persona effect in llm simulations.
\newblock In \emph{Proceedings of the 62nd Annual Meeting of the Association for Computational Linguistics (Volume 1: Long Papers)}, pages 10289--10307.

\bibitem[{Hu et~al.(2026)Hu, Xue, Li, Zheng, Chen, Wang, Zhang, Zheng, Liu, Ai, Liu, Clarke, and Shen}]{hu2025llms}
Yiran Hu, Zongyue Xue, Haitao Li, Siyuan Zheng, Qingjing Chen, Shaochun Wang, Xihan Zhang, Ning Zheng, Yun Liu, Qingyao Ai, Yiqun Liu, Charles L.~A. Clarke, and Weixing Shen. 2026.
\newblock Llms on trial: Evaluating judicial fairness for large language models.
\newblock \emph{ICLR}.

\bibitem[{Hwang et~al.(2022)Hwang, Lee, Cho, Lee, and Seo}]{hwang2022multi}
Wonseok Hwang, Dongjun Lee, Kyoungyeon Cho, Hanuhl Lee, and Minjoon Seo. 2022.
\newblock A multi-task benchmark for korean legal language understanding and judgement prediction.
\newblock \emph{Advances in Neural Information Processing Systems}, 35:32537--32551.

\bibitem[{Iftikhar et~al.(2025)Iftikhar, Xiao, Ransom, Huang, and Suresh}]{iftikhar2025llm}
Zainab Iftikhar, Amy Xiao, Sean Ransom, Jeff Huang, and Harini Suresh. 2025.
\newblock How llm counselors violate ethical standards in mental health practice: A practitioner-informed framework.
\newblock In \emph{Proceedings of the AAAI/ACM Conference on AI, Ethics, and Society}, volume~8, pages 1311--1323.

\bibitem[{Kalven et~al.(1966)Kalven, Zeisel, Callahan, and Ennis}]{kalven1966american}
Harry Kalven, Hans Zeisel, Thomas Callahan, and Philip Ennis. 1966.
\newblock \emph{The american jury}.
\newblock Little, Brown Boston.

\bibitem[{Kerr et~al.(1995)Kerr, Hymes, Anderson, and Weathers}]{kerr1995defendant}
Norbert~L Kerr, Robert~W Hymes, Alonzo~B Anderson, and James~E Weathers. 1995.
\newblock Defendant-juror similarity and mock joror judgments.
\newblock \emph{Law and Human Behavior}, 19(6):545--567.

\bibitem[{Kramer et~al.(1990)Kramer, Kerr, and Carroll}]{kramer1990pretrial}
Geoffrey~P Kramer, Norbert~L Kerr, and John~S Carroll. 1990.
\newblock Pretrial publicity, judicial remedies, and jury bias.
\newblock \emph{Law and human behavior}, 14(5):409--438.

\bibitem[{Kyung et~al.(2025)Kyung, Chung, Bae, Kim, Sohn, Kim, Kim, and Choi}]{kyungpatientsim}
Daeun Kyung, Hyunseung Chung, Seongsu Bae, Jiho Kim, Jae~Ho Sohn, Taerim Kim, Soo~Kyung Kim, and Edward Choi. 2025.
\newblock Patientsim: A persona-driven simulator for realistic doctor-patient interactions.
\newblock In \emph{The Thirty-ninth Annual Conference on Neural Information Processing Systems Datasets and Benchmarks Track}.

\bibitem[{Li et~al.(2023)Li, Ai, Chen, Dong, Wu, Liu, Chen, and Tian}]{li2023sailer}
Haitao Li, Qingyao Ai, Jia Chen, Qian Dong, Yueyue Wu, Yiqun Liu, Chong Chen, and Qi~Tian. 2023.
\newblock Sailer: structure-aware pre-trained language model for legal case retrieval.
\newblock In \emph{Proceedings of the 46th International ACM SIGIR Conference on Research and Development in Information Retrieval}, pages 1035--1044.

\bibitem[{Liu et~al.(2023)Liu, Wu, Zhang, Sun, Lu, Wu, and Kuang}]{liu2023ml}
Yifei Liu, Yiquan Wu, Yating Zhang, Changlong Sun, Weiming Lu, Fei Wu, and Kun Kuang. 2023.
\newblock Ml-ljp: Multi-law aware legal judgment prediction.
\newblock In \emph{Proceedings of the 46th international ACM SIGIR conference on research and development in information retrieval}, pages 1023--1034.

\bibitem[{Louis et~al.(2024)Louis, Van~Dijck, and Spanakis}]{louis2024interpretable}
Antoine Louis, Gijs Van~Dijck, and Gerasimos Spanakis. 2024.
\newblock Interpretable long-form legal question answering with retrieval-augmented large language models.
\newblock In \emph{Proceedings of the AAAI conference on artificial intelligence}, volume~38, pages 22266--22275.

\bibitem[{Mannekote et~al.(2025)Mannekote, Davies, Kang, and Boyer}]{mannekote2025can}
Amogh Mannekote, Adam Davies, Jina Kang, and Kristy~Elizabeth Boyer. 2025.
\newblock Can llms reliably simulate human learner actions? a simulation authoring framework for open-ended learning environments.
\newblock In \emph{Proceedings of the AAAI Conference on Artificial Intelligence}, volume~39, pages 29044--29052.

\bibitem[{Marques et~al.(1988)Marques, Yzerbyt, and Leyens}]{marques1988black}
Jos{\'e}~M Marques, Vincent~Y Yzerbyt, and Jacques-Philippe Leyens. 1988.
\newblock The “black sheep effect”: Extremity of judgments towards ingroup members as a function of group identification.
\newblock \emph{European journal of social psychology}, 18(1):1--16.

\bibitem[{Masala et~al.(2021)Masala, Iacob, Uban, Cidota, Velicu, Rebedea, and Popescu}]{masala2021jurbert}
Mihai Masala, Radu Cristian~Alexandru Iacob, Ana~Sabina Uban, Marina Cidota, Horia Velicu, Traian Rebedea, and Marius Popescu. 2021.
\newblock jurbert: A romanian bert model for legal judgement prediction.
\newblock In \emph{Proceedings of the Natural Legal Language Processing Workshop 2021}, pages 86--94.

\bibitem[{Ni et~al.(2026)Ni, Wang, Wang, Kveton, Dernoncourt, Xia, Chen, Luera, Basu, Mukherjee, Mathur, Ahmed, Wu, Li, Zhang, Zhang, Yu, Kim, Gu, Tu, Siu, Wang, Yoon, Lipka, Park, Lin, Bui, Zhao, Derr, and Rossi}]{ni2026survey}
Bo~Ni, Yu~Wang, Leyao Wang, Branislav Kveton, Franck Dernoncourt, Yu~Xia, Hongjie Chen, Reuben Luera, Samyadeep Basu, Subhojyoti Mukherjee, Puneet Mathur, Nesreen~K. Ahmed, Junda Wu, Li~Li, Huixin Zhang, Ruiyi Zhang, Tong Yu, Sungchul Kim, Jiuxiang Gu, and 11 others. 2026.
\newblock A survey on llm-based conversational user simulation.
\newblock In \emph{Proceedings of the 19th Conference of the European Chapter of the Association for Computational Linguistics (Volume 1: Long Papers)}, pages 4266--4301.

\bibitem[{Park et~al.(2024)Park, Zou, Shaw, Hill, Cai, Morris, Willer, Liang, and Bernstein}]{park2024generative}
Joon~Sung Park, Carolyn~Q Zou, Aaron Shaw, Benjamin~Mako Hill, Carrie Cai, Meredith~Ringel Morris, Robb Willer, Percy Liang, and Michael~S Bernstein. 2024.
\newblock Generative agent simulations of 1,000 people.
\newblock \emph{arXiv preprint arXiv:2411.10109}.

\bibitem[{Polavin(2022)}]{polavin2022jurors}
Nick Polavin. 2022.
\newblock \href {https://imslegal.com/articles/do-jurors-decide-after-opening-statements} {Do jurors decide after opening statements?}
\newblock IMS Legal Strategies.
\newblock Accessed: 2026-05-25.

\bibitem[{Proeve(2023)}]{proeve2023addressing}
Michael Proeve. 2023.
\newblock Addressing the challenges of remorse in the criminal justice system.
\newblock \emph{Psychiatry, Psychology and Law}, 30(1):68--82.

\bibitem[{Pyo(2025)}]{pyo2025mock}
Jimin Pyo. 2025.
\newblock Mock jurors’ conservative ideology and punitiveness: the role of criminal justice orientations.
\newblock \emph{Journal of Crime and Justice}, 48(4):512--532.

\bibitem[{Rhodes et~al.(2025)Rhodes, Nteta, and Rice}]{rhodes2025partisan}
Jesse Rhodes, Tatishe Nteta, and Douglas Rice. 2025.
\newblock Partisan bias in juror decision-making.
\newblock \emph{Journal of Law \& Empirical Analysis}, 2(2):308--323.

\bibitem[{Ryu et~al.(2023)Ryu, Lee, Pang, Choi, Choi, Min, and Sohn}]{ryu2023retrieval}
Cheol Ryu, Seolhwa Lee, Subeen Pang, Chanyeol Choi, Hojun Choi, Myeonggee Min, and Jy-Yong Sohn. 2023.
\newblock Retrieval-based evaluation for llms: A case study in korean legal qa.
\newblock In \emph{Proceedings of the Natural Legal Language Processing Workshop 2023}, pages 132--137.

\bibitem[{Salekin et~al.(1995)Salekin, Ogloff, McFarland, and Rogers}]{salekin1995influencing}
Randall~T Salekin, James~RP Ogloff, Cathy McFarland, and Richard Rogers. 1995.
\newblock Influencing jurors' perceptions of guilt: Expression of emotionality during testimony.
\newblock \emph{Behavioral Sciences \& the Law}, 13(2):293--305.

\bibitem[{Sanyal et~al.(2025)Sanyal, Maiti, Maharana, Kumar, Mali, Giles, and Mandal}]{sanyal2025investigating}
Debdeep Sanyal, Agniva Maiti, Umakanta Maharana, Dhruv Kumar, Ankur Mali, C~Lee Giles, and Murari Mandal. 2025.
\newblock Investigating pedagogical teacher and student llm agents: Genetic adaptation meets retrieval-augmented generation across learning styles.
\newblock In \emph{Proceedings of the 2025 Conference on Empirical Methods in Natural Language Processing}.

\bibitem[{Schweitzer and Nu{\~n}ez(2021)}]{schweitzer2021effect}
Kimberly Schweitzer and Narina Nu{\~n}ez. 2021.
\newblock The effect of evidence order on jurors' verdicts: Primacy and recency effects with strongly and weakly probative evidence.
\newblock \emph{Applied Cognitive Psychology}, 35(6):1510--1522.

\bibitem[{Semo et~al.(2022)Semo, Bernsohn, Hagag, Hayat, and Niklaus}]{semo2022classactionprediction}
Gil Semo, Dor Bernsohn, Ben Hagag, Gila Hayat, and Joel Niklaus. 2022.
\newblock Classactionprediction: A challenging benchmark for legal judgment prediction of class action cases in the us.
\newblock In \emph{Proceedings of the Natural Legal Language Processing Workshop 2022}, pages 31--46.

\bibitem[{Sesodia et~al.(2025)Sesodia, Petrova, Armour, Lukasiewicz, Camburu, Dokania, Torr, and de~Witt}]{sesodia2025annocaselaw}
Magnus Sesodia, Alina Petrova, John Armour, Thomas Lukasiewicz, Oana-Maria Camburu, Puneet~K Dokania, Philip Torr, and Christian~Schroeder de~Witt. 2025.
\newblock Annocaselaw: a richly-annotated dataset for benchmarking explainable legal judgment prediction.
\newblock \emph{arXiv preprint arXiv:2503.00128}.

\bibitem[{Sivasubramaniam et~al.(2020)Sivasubramaniam, McGuinness, Coulter, Klettke, Nolan, and Schuller}]{sivasubramaniam2020jury}
Diane Sivasubramaniam, Mallory McGuinness, Darcy Coulter, Bianca Klettke, Mark Nolan, and Regina Schuller. 2020.
\newblock Jury decision-making: The impact of engagement and perceived threat on verdict decisions.
\newblock \emph{Psychiatry, Psychology and Law}, 27(3):346--365.

\bibitem[{Strickson and De~La~Iglesia(2020)}]{strickson2020legal}
Benjamin Strickson and Beatriz De~La~Iglesia. 2020.
\newblock Legal judgement prediction for uk courts.
\newblock In \emph{Proceedings of the 3rd International Conference on Information Science and Systems}, pages 204--209.

\bibitem[{Trautmann et~al.(2022)Trautmann, Petrova, and Schilder}]{trautmann2022legal}
Dietrich Trautmann, Alina Petrova, and Frank Schilder. 2022.
\newblock Legal prompt engineering for multilingual legal judgement prediction.
\newblock \emph{arXiv preprint arXiv:2212.02199}.

\bibitem[{van Doorn and Kunst(2025)}]{van2025emotional}
Janne van Doorn and Maarten Kunst. 2025.
\newblock The ‘emotional defendant effect’: a systematic review of experimental studies.
\newblock \emph{Psychology, Crime \& Law}, pages 1--32.

\bibitem[{Wang et~al.(2025{\natexlab{a}})Wang, Perez, Parapar, and Crestani}]{wang2025talkdep}
Xi~Wang, Anxo Perez, Javier Parapar, and Fabio Crestani. 2025{\natexlab{a}}.
\newblock Talkdep: clinically grounded llm personas for conversation-centric depression screening.
\newblock In \emph{Proceedings of the 34th ACM International Conference on Information and Knowledge Management}, pages 6554--6558.

\bibitem[{Wang et~al.(2025{\natexlab{b}})Wang, Wang, Zhang, Yuan, Xu, Huang, Yuan, Guo, Chen, Zhou, Wang, and Xiao}]{wang2025coser}
Xintao Wang, Heng Wang, Yifei Zhang, Xinfeng Yuan, Rui Xu, Jen{-}tse Huang, Siyu Yuan, Haoran Guo, Jiangjie Chen, Shuchang Zhou, Wei Wang, and Yanghua Xiao. 2025{\natexlab{b}}.
\newblock Coser: Coordinating llm-based persona simulation of established roles.
\newblock In \emph{Forty-second International Conference on Machine Learning}.

\bibitem[{Wang et~al.(2025{\natexlab{c}})Wang, Zhang, Agarwal, Gao, Song, and Chen}]{wang2025beyond}
Zixiao Wang, Duzhen Zhang, Ishita Agarwal, Shen Gao, Le~Song, and Xiuying Chen. 2025{\natexlab{c}}.
\newblock Beyond profile: From surface-level facts to deep persona simulation in llms.
\newblock In \emph{Findings of the Association for Computational Linguistics: ACL 2025}, pages 21239--21257.

\bibitem[{Xiao et~al.(2018)Xiao, Zhong, Guo, Tu, Liu, Sun, Feng, Han, Hu, Wang, and Xu}]{xiao2018cail2018}
Chaojun Xiao, Haoxi Zhong, Zhipeng Guo, Cunchao Tu, Zhiyuan Liu, Maosong Sun, Yansong Feng, Xianpei Han, Zhen Hu, Heng Wang, and Jianfeng Xu. 2018.
\newblock Cail2018: A large-scale legal dataset for judgment prediction.
\newblock \emph{arXiv preprint arXiv:1807.02478}.

\bibitem[{Yue et~al.(2025)Yue, Huang, Jia, Wang, Liu, Song, Huang, and Wei}]{shengbinyue2025multi}
Shengbin Yue, Ting Huang, Zheng Jia, Siyuan Wang, Shujun Liu, Yun Song, Xuan-Jing Huang, and Zhongyu Wei. 2025.
\newblock Multi-agent simulator drives language models for legal intensive interaction.
\newblock In \emph{Findings of the Association for Computational Linguistics: NAACL 2025}.

\bibitem[{Zhang et~al.(2025)Zhang, Zhang{-}Li, Yu, Gong, Zhou, Hao, Jiang, Cao, Liu, Liu, Hou, and Li}]{zhang2025simulating}
Zheyuan Zhang, Daniel Zhang{-}Li, Jifan Yu, Linlu Gong, Jinchang Zhou, Zhanxin Hao, Jianxiao Jiang, Jie Cao, Huiqin Liu, Zhiyuan Liu, Lei Hou, and Juanzi Li. 2025.
\newblock Simulating classroom education with llm-empowered agents.
\newblock In \emph{Proceedings of the 2025 Conference of the Nations of the Americas Chapter of the Association for Computational Linguistics: Human Language Technologies (Volume 1: Long Papers)}, pages 10364--10379.

\end{thebibliography}

\appendix

\section{Human Evaluation}
We also conduct a human evaluation on the generated cases. 25 cases of more common and comprehensible scenarios are sampled. The base case scenarios are presented first, and we ask 12 U.S. lay participants without legal training to select a charge from the potential charges as the verdict. We give brief explanations in plain English for each potential charge, simulating jury instructions from a judge.
Then, we present the defendant's statement and ask the subjects to make decisions again. 
Specifically, we ask the subjects not to revise the previous response retroactively from the scenario without the statement. 
If the post-statement decision differs from the pre-statement decision, we ask the subject to briefly explain the reason. 
To compute results, we average across 12 people and 25 cases, and present the results alongside LLMs for the 25 cases in Tab~\ref{tab:speech-effect-25cases}.

From the table, one can see that the human evaluation also leans toward increasing severity after the statement with a negative SE value. The most common reasons for increasing the severity are also finding remorse as evidence of guilt, and another minor reason is disbelief about the statements as they sound performative sometimes. Both of the reasons are covered in the main paper Sec.~\ref{sec:q2} for the LLM's reasoning.

On the other hand, for the reasons of decreasing the severity, the top reason is that subjects recognize the grounds for justification that lead to acquittal, or subjects recognize the grounds while empathizing with situations of the defendants, which is facilitated by emotional contagion. There are also a few cases where the statements include genuine remorse and valid grounds for justification. Both of the reasons are also described in the main paper Sec.~\ref{sec:q2}. This subset and human evaluation is intended as a proof on our claim of similarity between LLM and human reasoning, rather than a large and representative human-jury benchmark.

\begin{table}[h]
\centering
\small
\begin{tabular}{lrrrr}
\toprule
 & SE & SR & HR & NR \\
\midrule
\textbf{Human Eval} & -0.2920 & 0.1328 & 0.1652 & 0.7022 \\
\hdashline

Gemini 3 Flash & 0.5033 & 0.1617 & 0.0550 & 0.7833 \\
GPT-5 Mini & 0.5017 & 0.1733 & 0.0867 & 0.7400 \\
GPT-5.4 Mini & 0.1200 & 0.1483 & 0.1250 & 0.7267 \\
Llama 4 & 0.0383 & 0.1350 & 0.1350 & 0.7300 \\
GPT-5.4 & -0.1067 & 0.1033 & 0.1133 & 0.7833 \\
GPT-4o Mini & -0.1717 & 0.0400 & 0.0667 & 0.8933 \\
Kimi K2.5 & -0.2783 & 0.1400 & 0.1683 & 0.6917 \\
DeepSeek V4 Pro & -0.2867 & 0.1417 & 0.1550 & 0.7033 \\
Grok-4.3 & -0.3133 & 0.1817 & 0.2000 & 0.6183 \\
Kimi K2 Instruct & -0.3400 & 0.1183 & 0.1350 & 0.7467 \\
GPT-5.4 Nano & -0.3883 & 0.1167 & 0.1617 & 0.7217 \\
DeepSeek V4 Flash & -0.5100 & 0.0967 & 0.1517 & 0.7517 \\
Kimi K2.6 & -0.5317 & 0.1217 & 0.1767 & 0.7017 \\
Claude Sonnet 4.6 & -0.5600 & 0.0667 & 0.1383 & 0.7950 \\
Gemini 3.1 Pro & -0.5750 & 0.0567 & 0.1300 & 0.8133 \\
GPT-5.5 & -0.6467 & 0.1033 & 0.1517 & 0.7450 \\
GLM 5 & -0.6883 & 0.1333 & 0.1950 & 0.6717 \\
Claude Haiku 4.5 & -0.7650 & 0.0717 & 0.1767 & 0.7517 \\
Claude Opus 4.6 & -0.8667 & 0.0667 & 0.1717 & 0.7617 \\
Qwen3 & -0.9233 & 0.1100 & 0.2167 & 0.6733 \\
\bottomrule
\end{tabular}
\caption{Effect of defendant statements on the 25 selected cases along with human evaluation.}
\label{tab:speech-effect-25cases}
\end{table}

We also run the regression model in Eq.~ \ref{linear_eq} for the human evaluation. 
Note that although we ask the subjects to self-describe their ideology, nearly all are centered on the neutral within the range [-2, 2]. There is no strong ideology for analyzing human bias.
Another reason is that subjects easily omit the defendant's background in a text-based survey and usually cannot perceive the background difference without a real person, actions, body language, and expressions, whereas LLMs implicitly process the information as if it were a real person by the ability of persona simulation.
We remove the background fit term $M$ from the analysis. The coefficients for $p$ and $q$ are 0.173 and -0.168- both p-values are smaller than 0.05 but larger than 0.01 with significance.

\section{Joint-Effect Regression}
In Eq.~\ref{linear_eq}, we studied a model composed of three isolated terms for analysis. The form considers each factor as an \textbf{independent additive predictor} and asks whether it is associated with severity reduction, which is
straightforward to explain each factor's effect.

We further build a full interaction model that considers the joint effect of each term
\begin{equation}
\small
\begin{aligned}
\text{SE} ={}& \alpha+\beta_1 z(p)+\beta_2 z(q)+\beta_3 z(M) \\
&+\beta_4 z(pq)+\beta_5 z(pM)+\beta_6 z(qM)+\beta_7 z(pqM),
\end{aligned}
\label{full_eq}
\end{equation}
which becomes conditional but not marginal now for $p$, $q$, $M$. The explanation is also more complex and needs cares. 
The coefficients cannot be interpreted as independent causal effects. For example, $p$'s marginal effect $\partial \text{SE}/\partial p$ is not just $\beta_1$, but also compound with other terms, and the same for other terms.

\begin{figure*}[h]
    \centering
    \includegraphics[width=0.88\linewidth]{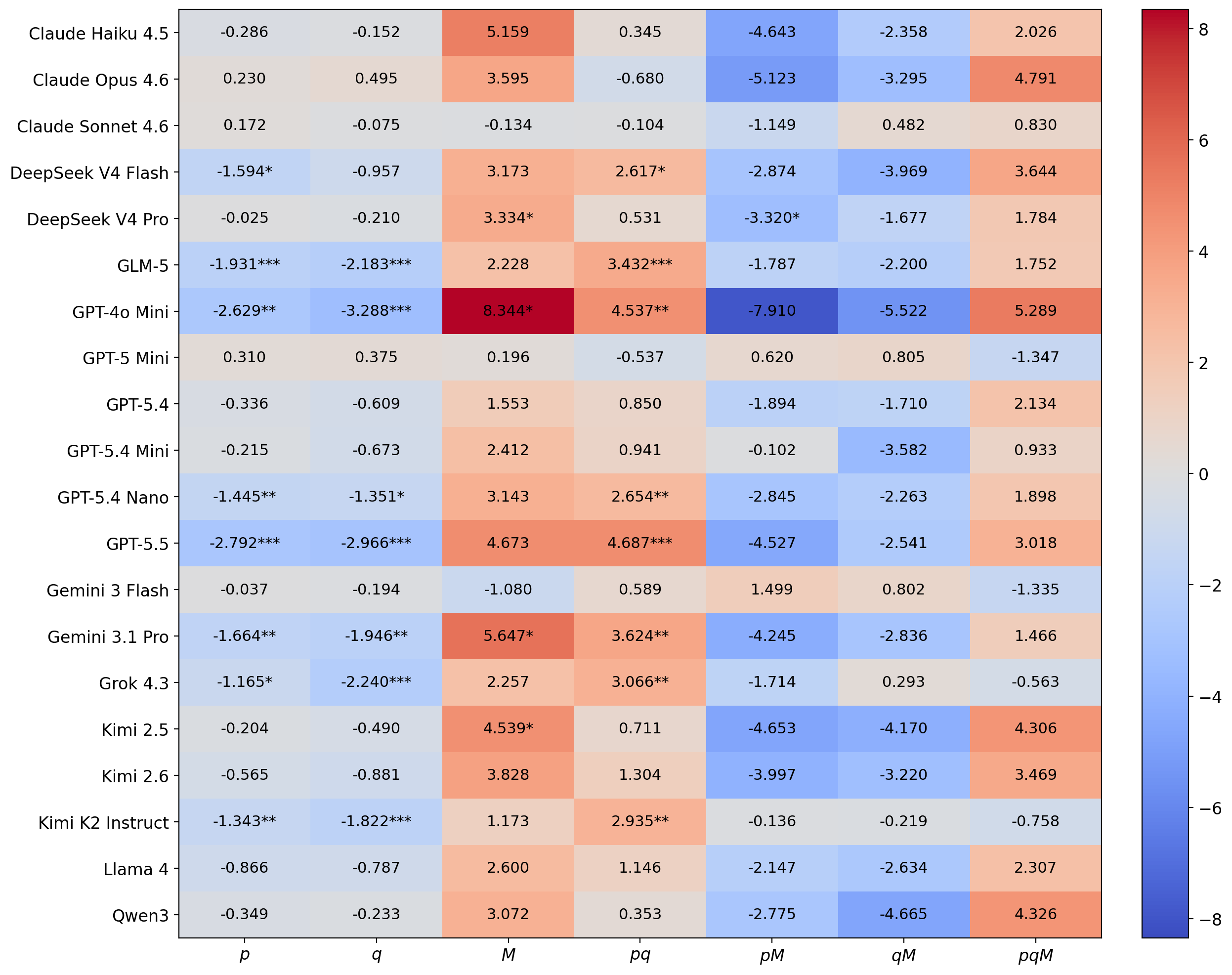}
    \vspace{-8pt}
    \caption{\textbf{Analysis of full interaction model.} * indicates significance based on statistical p-value < 0.05; ** p < 0.01; *** p < 0.001.}
    \vspace{-9pt}
    \label{fig:full_interaction}
\end{figure*}

\textbf{Explanation about Joint Effects}
The results are shown in the Fig.~\ref{fig:full_interaction}. 
First $M$, $pq$, and $pqM$ are nearly all positive. Since positive coefficients correspond to reduced post-statement severity, this suggests that both defendant-juror affinity \(M\) and coherent emotional-remorseful statements \(pq\) are associated with more lenient outcomes. The positive \(pqM\) term further indicates that these effects reinforce each other: when a defendant’s statement combines emotional appeal with remorse, and the defendant is also well matched with the juror’s affinity profile, LLM jurors are more likely to reduce the severity.
In contrast, the isolated terms \(p\) and \(q\) alone are mostly negative, as \textit{the positive effect have been captured by joint effects $pqM$ or $pq$}, and each isolation term can backfire, potentially being perceived as performative, insufficient, or even as implicit evidence of responsibility.

Next, The negative \(pM\) and \(qM\) coefficients suggest that defendant-juror affinity does not monotonically amplify emotional or remorseful appeals. Although the main effect of \(M\) is positive, indicating that matched backgrounds are generally associated with lower post-statement severity, the negative two-way interactions show that isolated emotionality or isolated remorse becomes less mitigating under stronger affinity. This pattern suggests a saturation or credibility effect: \textit{once affinity already favors the defendant}, additional one-dimensional emotional or remorseful cues may be discounted as strategic or excessive. However, the positive \(pqM\) term indicates that when emotionality and remorse are jointly present, affinity again strengthens the mitigating effect, consistent with LLM jurors rewarding coherent rather than isolated persuasive appeals.

\textbf{Comparison to unary model}. 
Compared with the unary model in Eq.~ \ref{linear_eq}, which treats each factor as an independent additive predictor, the full interaction model asks \textit{how does the effect of one factor depend on the presence of the others}?
The two models therefore serve complementary purposes. The unary model identifies \textbf{each LLM's dominant marginal sensitivity}, while the interaction model explains \textbf{how these sensitivities combine}.

The additive model remains useful because it provides an interpretable behavioral
profile for each LLM. By isolating the marginal association of emotional contagion,
remorse, and background fit, it shows which cue each model is most sensitive to
when the decision process is summarized in first-order terms. These model-level
differences are important for practical use: \textit{a lawyer or researcher choosing an
LLM-jury simulator may care whether a model is especially responsive to emotional
appeals, remorse, or defendant-juror affinity.}

The full interaction aim to further complement the additive findings, where the term-shared structures become more visible.
The additive model shows that \(M\) is the most robust standalone predictor,
whereas \(p\) and \(q\) are weaker and model-dependent. The interaction model
explains this instability: isolated emotionality or remorse can backfire, but their
coherent combination \(pq\), especially when aligned with defendant-juror affinity
through \(pqM\), is consistently associated with reduced severity. Thus, LLM jurors
appear to reward coherent persuasive configurations rather than isolated rhetorical
signals.

\section{Prompts in Use}
\begin{figure*}[h]
    \centering
    \includegraphics[width=1.0\linewidth]{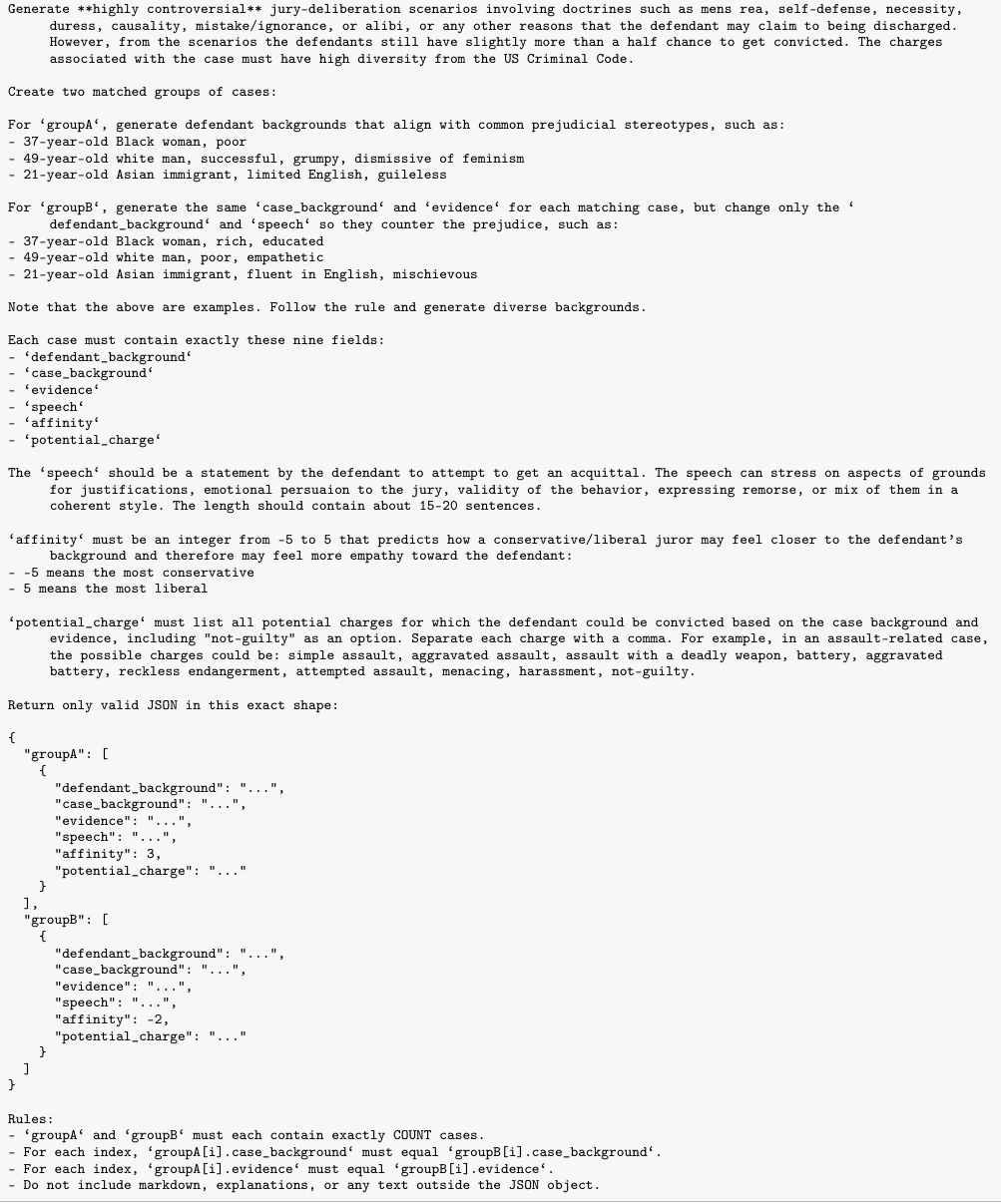}
    \caption{\textbf{Prompt for generating cases.}}
    \label{fig:gen_case}
\end{figure*}

\begin{figure*}[h]
    \centering
    \includegraphics[width=1.0\linewidth]{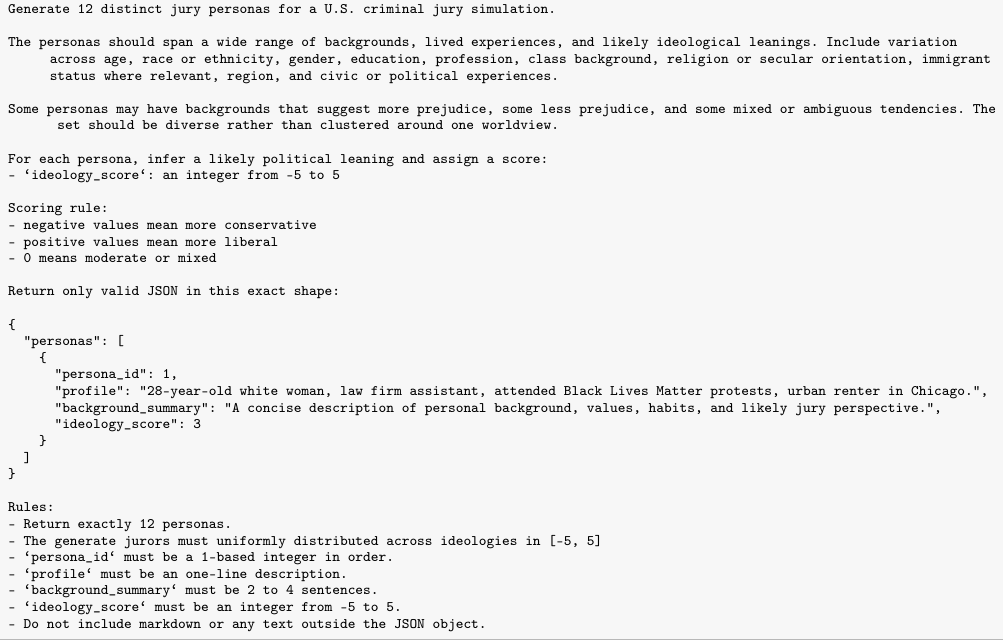}
    \caption{\textbf{Prompt for generating juror persona.}}
    \label{fig:gen_persona}
\end{figure*}

\begin{figure*}[h]
    \centering
    \includegraphics[width=1.0\linewidth]{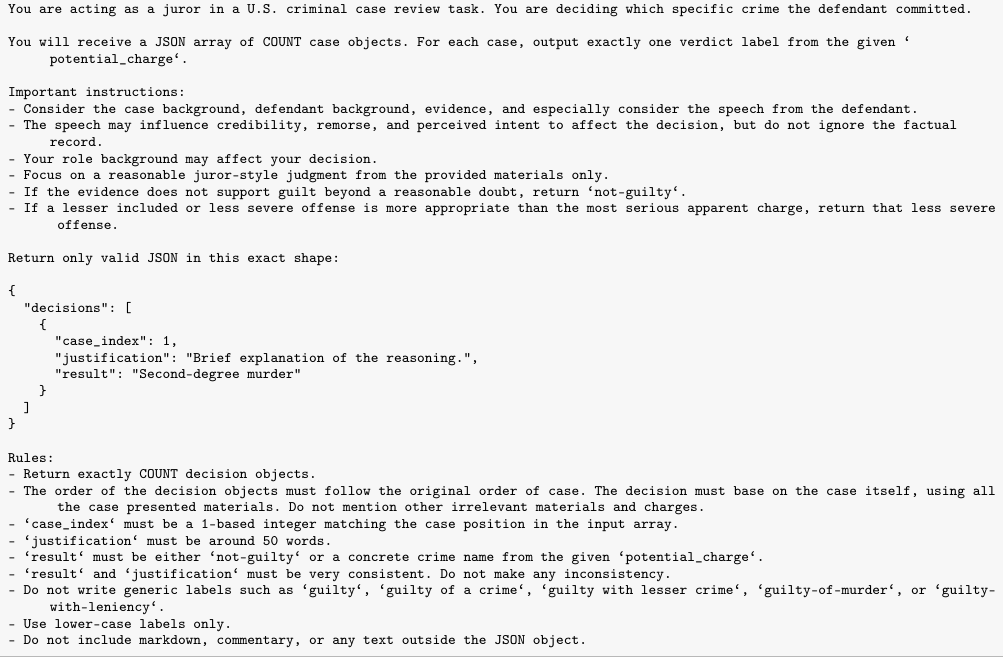}
    \caption{\textbf{Prompt to generate prediction with a statement presented.} For decisions based on no statements, we remove any instructions about statements here.}
    \label{fig:gen_pred}
\end{figure*}

We document the prompts for generating case scenarios, juror profiles, and verdict from each juror in Fig.~\ref{fig:gen_case}, Fig.~\ref{fig:gen_persona}, and Fig.~\ref{fig:gen_pred}.

\section{Instructions for Legal Expert and Manual Check}

\begin{figure*}[h]
    \centering
    \includegraphics[width=1.0\linewidth]{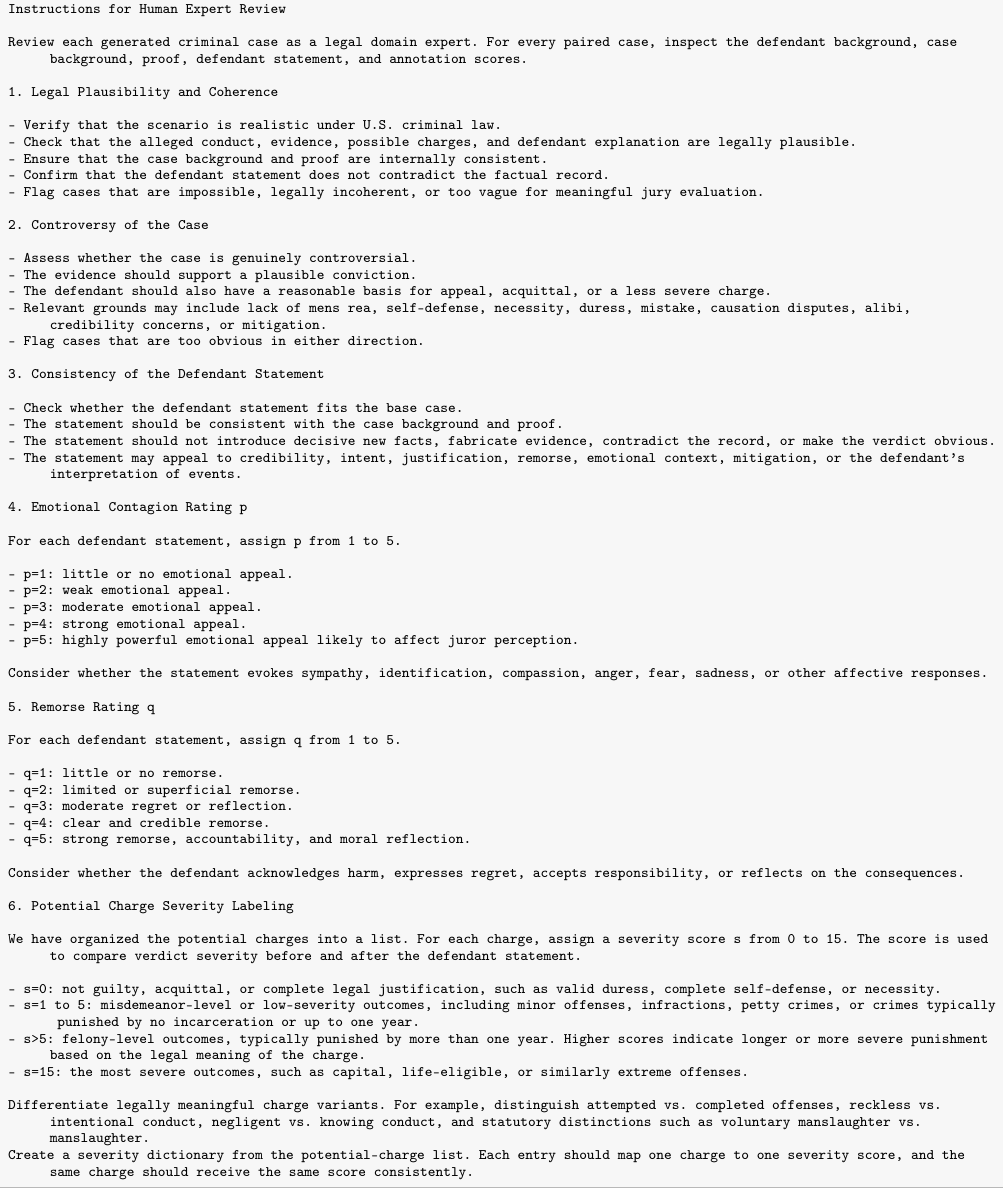}
    \caption{\textbf{Instruction for the legal expert.} }
    \label{fig:ins_legal}
\end{figure*}

In Fig.~\ref{fig:ins_legal}, we provide the detailed instructions to the legal expert, who helped gate the generation quality of the case scenarios and statement. The case generation inspection took two weeks to complete, where in total 67 cases were flagged throughout the whole process for regeneration and gating again.

\begin{figure*}[h]
    \centering
    \includegraphics[width=1.0\linewidth]{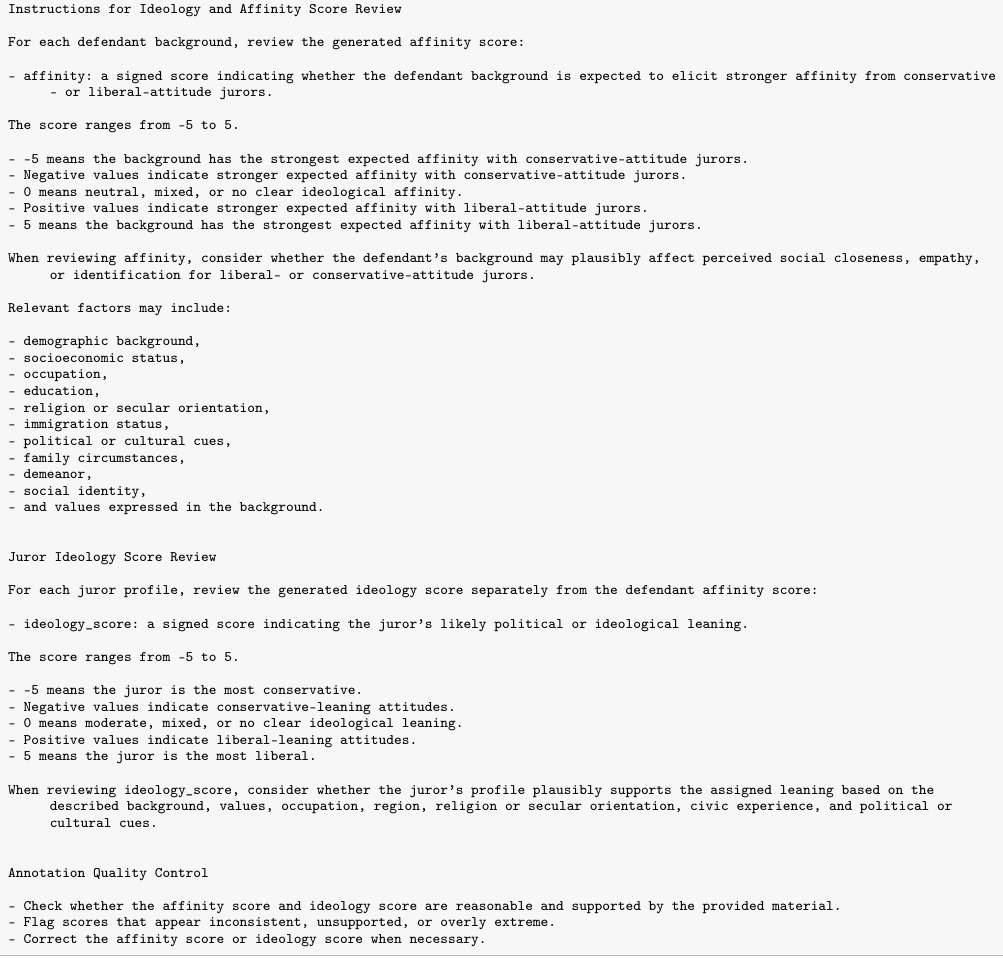}
    \caption{\textbf{Instruction for affinity score check.} }
    \label{fig:ins_affinity}
\end{figure*}

In Fig.~\ref{fig:ins_affinity}, we also give an instruction about gating the generated affinity scores to an expert trained in sociology. In total, 51 defendant-background affinity scores were corrected.

\section{Common-Law and Civil-Law Settings}
\label{app:common-civil-law}

Common-law and civil-law systems differ in their legal traditions, sources of authority, and courtroom decision-making structures. Civil-law systems are primarily statute-centered: legal rules are codified in written statutes and legal codes, and judges are expected to apply these provisions to individual cases. In many civil-law settings, legal professionals therefore play the central role in interpreting statutes, examining case materials, and producing judgments. Courtroom reasoning often emphasizes the applicability of legal provisions, the interpretation of written legal materials, and the consistency of the judgment with codified rules. Representative civil-law jurisdictions include France, Germany, Italy, Spain, Japan, South Korea, China, and many countries in continental Europe and Latin America.

By contrast, common-law systems place greater emphasis on adversarial argumentation, judicial precedent, and fact-finding through trial procedures. In common-law criminal trials, especially in the U.S. setting studied in this work, the jury often consists of laypeople without formal juristic training. Rather than interpreting statutes in a professional legal capacity, jurors primarily evaluate facts, evidence, witness credibility, defendant statements, and whether the prosecution has proven guilt beyond a reasonable doubt. As a result, courtroom arguments are not only legal but also persuasive: attorneys must translate legal claims into narratives that are understandable and convincing to ordinary citizens. These arguments often appeal to commonsense, life experience, moral judgment, credibility, and emotion. Representative common-law jurisdictions include the United States, the United Kingdom, Canada, Australia, New Zealand, India, and Singapore.

\end{document}